# Testing of Detection Tools for AI-Generated Text

Debora Weber-Wulff (University of Applied Sciences HTW Berlin, Germany, weberwu@htw-berlin.de), (corresponding author)
Alla Anohina-Naumeca (Riga Technical University, Latvia, alla.anohina-naumeca@rtu.lv)
Sonja Bjelobaba (Uppsala University, Sweden, sonja.bjelobaba@crb.uu.se),
Tomáš Foltýnek (Masaryk University, Czechia, foltynek@fi.muni.cz)
Jean Guerrero-Dib (Universidad de Monterrey, Mexico, jean.guerrero@udem.edu.mx),
Olumide Popoola (Queen Mary University of London, UK, O.Popoola@qmul.ac.uk),
Petr Šigut (Masaryk University, Czechia, petrsigut10@gmail.com),
Lorna Waddington (University of Leeds, UK, l.l.waddington@leeds.ac.uk)

## Abstract

Recent advances in generative pre-trained transformer large language models have emphasised the potential risks of unfair use of artificial intelligence (AI) generated content in an academic environment and intensified efforts in searching for solutions to detect such content. The paper examines the general functionality of detection tools for artificial intelligence generated text and evaluates them based on accuracy and error type analysis. Specifically, the study seeks to answer research questions about whether existing detection tools can reliably differentiate between human-written text and ChatGPT-generated text, and whether machine translation and content obfuscation techniques affect the detection of AI-generated text. The research covers 12 publicly available tools and two commercial systems (Turnitin and PlagiarismCheck) that are widely used in the academic setting. The researchers conclude that the available detection tools are neither accurate nor reliable and have a main bias towards classifying the output as human-written rather than detecting AI-generated text. Furthermore, content obfuscation techniques significantly worsen the performance of tools. The study makes several significant contributions. First, it summarises up-to-date similar scientific and non-scientific efforts in the field. Second, it presents the result of one of the most comprehensive tests conducted so far, based on a rigorous research methodology, an original document set, and a broad coverage of tools. Third, it discusses the implications and drawbacks of using detection tools for AI-generated text in academic settings.

# 1. Introduction

Higher education institutions (HEIs) play a fundamental role in society. They shape the next generation of professionals through education and skill development, simultaneously providing hubs for research, innovation, collaboration with business, and civic engagement. It is also in higher education that students form and further develop their personal and professional ethics and values. Hence, it is crucial to uphold the integrity of the assessments and diplomas provided in tertiary education.

The introduction of unauthorised content generation—"the production of academic work, in whole or part, for academic credit, progression or award, whether or not a payment or other favour is involved, using unapproved or undeclared human or technological assistance" (Foltýnek et al., 2023)—into higher education contexts poses potential threats to academic integrity. Academic integrity is understood as "compliance with ethical and professional principles, standards and practices by individuals or institutions in education, research and scholarship" (Tauginienė et al., 2018).

Recent advancements in artificial intelligence (AI), particularly in the area of the generative pre-trained transformer (GPT) large language models (LLM), have led to a range of publicly available online text generation tools. As these models are trained on human-written texts, the content generated by these tools can be quite difficult to distinguish from human-written content. They can thus be used to complete assessment tasks at HEIs.

Despite the fact that unauthorised content generation created by humans, such as contract cheating (Clarke & Lancaster, 2006), has been a well-researched form of student cheating for almost two decades now, HEIs were not prepared for such radical improvements in automated tools that make unauthorised content generation so easily accessible for students and researchers. The availability of tools based on GPT-3 and newer LLMs, ChatGPT (OpenAI, 2023, 13 February) in particular, as well as other types of AI-based tools such as machine translation tools or image generators, have raised many concerns about how to make sure that no academic performance deception attempts have been made. The availability of ChatGPT has forced HEIs into action.

Unlike contract cheating, the use of AI tools is not automatically unethical. On the contrary, as AI will permeate society and most professions in the near future, there is a need to discuss with students the benefits and limitations of AI tools, provide them with opportunities to expand their knowledge of such tools, and teach them how to use AI ethically and transparently.

Nonetheless, some educational institutions have directly prohibited the use of ChatGPT (Johnson, 2023), and others have even blocked access from their university networks (Elsen-Rooney, 2023), although this is just a symbolic measure with virtual private networks quite prevalent. Some conferences have explicitly prohibited AI-generated content in conference submissions, including machine-learning conferences (ICML, 2023). More recently, Italy became the first country in the world to ban the use of ChatGPT, although that decision has in the meantime been rescinded (Schechner, 2023). Restricting the use of AI-generated content has naturally led to the desire for simple detection tools. Many free online tools that claim to be able to detect AI-generated text are already available.

Some companies do urge caution when using their tools for detecting AI-generated text for taking punitive measures based solely on the results they provide. They acknowledge the limitations of their tools, e.g. OpenAI explains that there are several ways to deceive the tool (OpenAI, 2023, 31 January). Turnitin made a guide for teachers on how they should approach the students whose work was flagged as AI-generated (Turnitin, 2023, 16 March). Nevertheless, four different companies (GoWinston, 2023; Content at Scale, 2023; Compilatio, 2023; GPTZero, 2023) claim to be the best on the market.

The aim of this paper is to examine the general functionality of tools for the detection of the use of ChatGPT in text production, assess the accuracy of the output provided by these tools, and their efficacy in the face of the use of obfuscation techniques such as online paraphrasing tools, as well as the influence of machine translation tools to human-written text.

Specifically, the paper aims to answer the following research questions:

**RQ1:** Can detection tools for AI-generated text reliably detect human-written text?
**RQ2:** Can detection tools for AI-generated text reliably detect ChatGPT-generated text?
**RQ3:** Does machine translation affect the detection of human-written text?
**RQ4:** Does manual editing or machine paraphrasing affect the detection of ChatGPT-generated text?
**RQ5:** How consistent are the results obtained by different detection tools for AI-generated text?

The next section briefly describes the concept and history of LLMs. It is followed by a review of scientific and non-scientific related work and a detailed description of the research methodology. After that, the results are presented in terms of accuracy, error analysis, and usability issues. The paper ends with discussion points and conclusions made.

# 2. Large Language Models

We understand LLMs as systems trained to predict the likelihood of a specific character, word, or string (called a *token*) in a particular context (Bender et al., 2021). Such statistical language models have been used since the 1980s (Rosenfeld, 2000), amongst other things for machine translation and automatic speech recognition. Efficient methods for the estimation of word representations in multidimensional vector spaces (Mikolov et al., 2013), together with the attention mechanism and transformer architecture (Vaswani et al., 2017) made generating human-like text not only possible, but also computationally feasible.

ChatGPT is a Natural Language Processing system that is owned and developed by OpenAI, a research and development company established in 2015. Based on the transformer architecture, OpenAI released the first version of GPT in June 2018. Within less than a year, this version was replaced by a much improved GPT-2, and then in 2020 by GPT-3. (Marr, 2023). This version could generate coherent text within a given context. This was in many ways a game-changer, as it is capable of creating responses that are hard to distinguish from human-written text (Borji, 2023; Brown et al., 2020). As 7% of the training data is on languages other than English, GPT-3 can also perform multilingually (Brown et al., 2020). In November 2022, ChatGPT was launched. It demonstrated significant improvements in its capabilities, a user-friendly interface, and it was widely reported in the general press. Within two months of its launch, it had over 100 million subscribers and was labelled "the fastest growing consumer app ever" (Milmo, 2023).

AI in education brings both challenges and opportunities. Authorised and properly acknowledged usage of AI tools, including LLMs, is not per se a form of misconduct (Foltýnek et al., 2023). However, using AI tools in an educational context for unauthorised content generation (Foltýnek et al., 2023) is indeed a form of academic misconduct (Tauginienė et al., 2018). Although LLMs have become known to the wider public after the release of ChatGPT, there is no reason to assume that they have not been used to create unauthorised and undeclared content even before that date. The accessibility, quantity, and recent development of AI tools have led many educators to demand technical solutions to help them distinguish between human-written and AI-generated texts.

For more than two decades, educators have been using software tools in an attempt to detect academic misconduct. This includes using search engines and text-matching software in order to detect instances of potential plagiarism. Although such automated detection can identify some plagiarism, previous research by Foltýnek et al. (2020) has shown that text-matching software not only do not find all plagiarism, but furthermore will also mark non-plagiarised content as plagiarism, thus providing false positive results. This is a worst-case scenario in academic settings, as an honest student can be accused of misconduct. In order to avoid such a scenario, now, when the market has responded with the introduction of dozens of tools for AI-generated text, it is important to discuss whether these tools clearly distinguish between human-written and machine-generated content.

# 3. Related work

The development of LLMs has led to an acceleration of different types of efforts in the field of automatic detection of AI-generated text. Firstly, several researchers has studied human abilities to detect machine-generated texts (e.g. Guo et al., 2023; Ippolito et al., 2020; Ma et al., 2023). Secondly, some attempts have been made to build benchmark text corpora to detect AI-generated texts effectively; for example, Liyanage et al. (2022) have offered synthetic and partial text substitution datasets for the academic domain. Thirdly, many research works are focused on developing new or fine-tuning parameters of the already pre-trained models of machine-generated text (e.g. Chakraborty et al., 2023; Devlin et al., 2019).

These efforts provide a valuable contribution to improving the performance and capabilities of detection tools for AI-generated text. In this section, the authors of the paper mainly focus on studies that compare or test the existing detection tools that educators can use to check the originality of students' assignments. The related works examined in the paper are summarised in Table 1. They are categorised as published scientific publications, preprints and non-scientific publications. It is worth mentioning that although there are many non-scientific comparisons on the Internet made by individuals and organisations, this table includes only those with the higher coverage of tools and/or at least partly described methodology of experiments.

Table 1. Related work

| Source | Detection tools used | Dataset | Evaluation metrics |
|---|---|---|---|
| **Published scientific publications** | | | |
| (Aydin & Karaarslan, 2022) | 1<br>iThenticate | An article with three sections: the text written by the paper's authors, the ChatGPT -paraphrased abstract text of articles, the content generated by ChatGPT answering specific questions | N/A |
| (Anderson et al., 2023) | 1<br>(GPT-2 Output Detector) | Two ChatGPT-generated essays and the same essays paraphrased by AI | N/A |
| **Preprints** | | | |
| (Gao et al., 2022) | 2<br>(Plagiarismdetector.net, GPT-2 Output Detector) | 50 ChatGPT-generated scientific abstracts | AUROC |
| (Khalil & Er, 2023) | 3<br>(iThenticate, Turnitin, ChatGPT) | 50 essays generated by ChatGPT on various topics (such as physics laws, data mining, global warming, driving schools, machine learning, etc.) | True positive,<br>False negative |
| (Wang et al., 2023) | 6 (GPT2-Detector, RoBERTa-QA, DetectGPT, GPTZero Writer, OpenAI Text Classifier) | • Q&A-GPT: 115K pairs of human-generated answers (taken from Stack Overflow) and ChatGPT generated answers (for the same topic) for 115K questions<br>• Code2Doc-GPT: 126K samples from CodeSearchNet and GPT code description for 6 programming languages<br>• 226.5K pairs of code samples human and ChatGPT generated (APPS-GPT, CONCODE-GPT, Doc2Code-GPT)<br>• Wiki-GPT dataset: 25K samples of human-generated and GPT polished texts | AUC scores, False positive rate, False negative rate |
| (Pegoraro et al., 2023) | 24 approaches and tools (among them online tools ZeroGPT, OpenAI Text Classifier, GPTZero, Hugging Face, Writefull, Copyleaks, Content at Scale, Originality.ai, Writer, Draft and Goal) | 58,546 responses generated by humans and 72,966 responses generated by the ChatGPT model, resulting in 131,512 unique samples that address 24,322 distinct questions from various fields, including medicine, opendomain, and finance | True positive rate, True negative rate |
| **Non-scientific publications** | | | |
| (Gewirtz, 2023) | 3 (GPT-2 Output Detector, Writer, Content at Scale) | • 3 human-generated texts<br>• 3 ChatGPT-generated texts | N/A |
| (van Oijen, 2023) | 7 (Content at Scale, Copyleaks, Corrector App, Crossplag, GPTZero, OpenAI, Writer) | • 10 generated passages based on prompts (factual info, rewrites of existing test, fictional scenarios, advice, explanations at different levels, impersonation of a specified character, Dutch translation)<br>• 5 human-generated text from different sources (Wikipedia, SURF, Alice in Wonderland, Reddit post) | Accuracy |

| (Compilatio, 2023) | 11 (Compilatio, Draft and Goal, GLTR, GPTZero, Content at Scale, DetectGPT, Crossplag, Kazan SEO, AI Text Classifier, Copyleaks, Writer AI Content Detector) | ● 50 human-written texts<br>● 75 texts generated by ChatGPT and YouChat | Reliability (the number of correctly classified/the total number of text passages) |
|---|---|---|---|
| (Demers, 2023) | 16 (Originality AI, Writer, Copyleaks, Open AI Text Classifier, Crossplag, GPTZero, Sapling, Content At Scale, ZeroGPT, GLTR, Hugging Face, Corrector, Writeful, Hive Moderation, Paraphrasing tool AI Content Detector, AI Writing Check) | ● Human writing sample<br>● ChatGPT 4 writing sample<br>● ChatGPT 4 writing sample with the additional prompt "beat detection" | N/A |

Some researchers have used known text-matching software to check if they are able to find instances of plagiarism in the AI-generated text. Aydin and Karaarslan (2022) tested the iThenticate system and have revealed that the tool has found matches with other information sources both for ChatGPT-paraphrased text and -generated text. They also found that ChatGPT does not produce original texts after paraphrasing, as the match rates for paraphrased texts were very high in comparison to human-written and ChatGPT-generated text passages. In the experiment of Gao et al. (2022), Plagiarismdetector.net recognized nearly all of the fifty scientific abstracts generated by ChatGPT as completely original.

Khalil and Er (2022) fed 50 ChatGPT-generated essays into two text-matching software systems (25 essays to iThenticate and 25 essays to the Turnitin system), although they are just different interfaces to the same engine. They found that 40 (80%) of them were considered to have a high level of originality, although they defined this as a similarity score of 20% or less. Khalil and Er (2022) also attempted to test the capabilities of ChatGPT to detect if the essays were generated by ChatGPT and state an accuracy of 92%, as 46 essays were supposedly said to be cases of plagiarism.

The authors of this paper consider the study of Khalil and Er (2022) to be problematic for two reasons. First, it is worth noting that the application of text-matching software systems to the detection of LLM-generated text makes little sense because of the stochastic nature of the word selection. Second, since an LLM will "hallucinate", that is, make up results, it cannot be asked whether it is the author of a text.

As of May 2023, ChatGPT now issues a warning to such questions such as: "As an AI language model, I cannot verify the specific source or origin of the paragraph you provided."

Several researchers focused on testing sets of free and/or paid detection tools for AI-generated text. Wang et al. (2023) checked the performance of detection tools on both natural language content and programming code and determined that “detecting ChatGPT-generated code is even more difficult than detecting natural language contents.” They also state that tools often exhibit bias, as some of them have a tendency to predict that content is ChatGPT generated (positive results), while others tend to predict that it is human-written (negative results).

By testing fifty ChatGPT-generated paper abstracts on the GPT-2 Output detector, Gao et al. (2022) concluded that the detector was able to make an excellent distinction between original and generated abstracts because the majority of the original abstracts were scored extremely low (corresponding to human-written content) while the detector found a high probability of AI-generated text in the majority (33 abstracts) of the ChatGPT-generated abstracts with 17 abstracts scored below 50%.

Pegoraro et al. (2023) tested not only online detection tools for AI-generated text but also many of the existing detection approaches and claimed that detection of the ChatGPT-generated text passages is still a very challenging task as the most effective online detection tool can only achieve a success rate of less than 50%. They also concluded that most of the analysed tools tend to classify any text as human-written.

Tests completed by van Oijen (2023) showed that the overall accuracy of tools in detecting AI-generated text reached only 27.9%, and the best tool achieved a maximum of 50% accuracy, while the tools reached an accuracy of almost 83% in detecting human-written content. The author concluded that detection tools for AI-generated text are “no better than random classifiers” (van Oijen, 2023). Moreover, the tests provided some interesting findings; for example, the tools found it challenging to detect a piece of human-written text that was rewritten by ChatGPT or a text passage that was written in a specific style. Additionally, there was not a single attribution of a human-written text to AI-generated text, that is, an absence of false positives.

Although Demers (2023) only provided results of testing without any further analysis, their examination allows making conclusions that a text passage written by a human was recognised as human-written by all tools, while ChatGPT-generated text had a mixed evaluation with the tendency to be predicted as human-written (10 tools out of 16) that increased even further for the ChatGPT writing sample with the additional prompt “beat detection” (12 tools out of 16).

In the tests conducted by Compilatio, the detection tools for AI-generated text detected human-written text with reliability in the range of 78–98% and AI-generated text in the range of 56–88%. Gewirtz’ (2023) results on testing three human-written and three ChatGPT-generated texts demonstrated that two of the selected detection tools for AI-generated text could reach only 50% accuracy and one an accuracy of 66%.

The effect of paraphrasing on the performance of detection tools for AI-generated text has also been studied. For example, Anderson et al. (2023) concluded that paraphrasing has significantly lowered the detection capabilities of the GPT-2 Output Detector by increasing the score for human-written content from 0.02% to 99.52% for the first essay and from 61.96% to 99.98% for the second essay. Krishna et al. (2023) applied paraphrasing to the AI-generated texts and revealed that it significantly lowered the detection accuracy of five detection tools for AI-generated text used in the experiments.

The results of the above-mentioned studies suggest that detecting AI-generated text passages is still challenging for existent detection tools for AI-generated text, whereas human-written texts are usually identified quite accurately (accuracy above 80%). However, the ability of tools to identify AI-generated text is under question as their accuracy in many studies was only around 50% or slightly above. Depending on the tool, a bias may be observed identifying a piece of text as either ChatGPT-generated or human-written. In addition, tools have difficulty identifying the source of the text if ChatGPT transforms human-written text or generates text in a particular style (e.g. a child's explanation). Furthermore, the performance of detection tools significantly decreases when texts are deliberately modified by paraphrasing or re-writing. Detection of the AI-generated text remains challenging for existing detection tools, but detecting ChatGPT-generated code is even more difficult.

Existing research has several shortcomings:

- quite often experiments are carried out with a limited number of detection tools for AI-generated text on a limited set of data;
- sometimes human-written texts are taken from publicly available websites or recognised print sources, and thus could potentially have been previously used to train language models and/or provide no guarantee that they were actually written by humans;
- the methodological aspects of the research are not always described in detail and are thus not available for replication;
- testing whether the AI-generated and further translated text can influence the accuracy of the detection tools is not discussed at all;
- a limited number of measurable metrics is used to evaluate the performance of detection tools, ignoring the qualitative analysis of results, for example, types of classification errors that can have significant consequences in an academic setting.

# 4. Methodology

## 4.1 Test cases

The focus of this investigation is determining the accuracy of tools which state that they are able to detect AI-generated text. In order to do so, a number of situational parameters were set up for creating the test cases for the following categories of English-language documents:

- human-written;
- human-written in a non-English language with a subsequent AI/machine translation to English;

- AI-generated text;
- AI-generated text with subsequent human manual edits;
- AI-generated text with subsequent AI/machine paraphrase.

For the first category (called 01-Hum), the specification was made that 10.000 characters (including spaces) were to be written at about the level of an undergraduate in the field of the researcher writing the paper. These fields include academic integrity, civil engineering, computer science, economics, history, linguistics, and literature. None of the text may have been exposed to the Internet at any time or even sent as an attachment to an email. This is crucial because any material that is on the Internet is potentially included in the training data for an LLM.

For the second category (called 02-MT), around 10.000 characters (including spaces) were written in Bosnian, Czech, German, Latvian, Slovak, Spanish, and Swedish. None of this texts may have been exposed to the Internet before, as for 01-Hum. Depending on the language, either the AI translation tool DeepL (3 cases) or Google Translate (6 cases) was used to produce the test documents in English.

It was decided to use ChatGPT as the only AI-text generator for this investigation, as it was the one with the largest media attention at the beginning of the investigation. Each researcher generated two documents with the tool using different prompts, (03-AI and 04-AI) with a minimum of 2000 characters each and recorded the prompts. The language model from February 13, 2023 was used for all test cases.

Two additional texts of at least 2000 characters were generated using fresh prompts for ChatGPT, then the output was manipulated. It was decided to use this type of test case, as students will have a tendency to obfuscate results with the expressed purpose of hiding their use of an AI-content generator. One set (05-ManEd) was edited manually with a human exchanging some words with synonyms or reordering sentence parts and the other (06-Para) was rewritten automatically with the AI-based tool Quillbot (Quillbot, 2023), using the default values of the tool for modes (Standard) and synonym level. Documentation of the obfuscation, highlighting the differences between the texts, can be found in the Appendix.

With nine researchers preparing texts (the eight authors and one collaborator), 54 test cases were thus available for which the ground truth is known.

## 4.2 AI-Generated Text Detection Tool Selection

A list of detection tools for AI-generated text was prepared using social media and Google search. Overall, 18 tools were considered, out of which 6 were excluded: 2 were not available, 2 were not online applications but Chrome extensions and thus out of the scope of this investigation, 1 required payment, and 1 did not produce any quantifiable result.

The company Turnitin approached the research group and offered a login, noting that they could only offer access from early April 2023. It was decided to test the system, although it is not free, because it is so widely used and already widely discussed in academia. Another company, PlagiarismCheck, was also advertising that it had a detection tool for AI-generated text in addition to its text-matching detection system. It was decided to ask them if they wanted to be part of the test as well, as the researchers did not want to have only one paid system. They agreed and provided a login in early May. We caution that their results may be different from the free tools used, as the companies knew that the submitted documents were part of a test suite and they were able to use the entire test document.

The following 14 detection tools were tested:

- Check For AI (https://checkforai.com)
- Compilatio (https://ai-detector.compilatio.net/)
- Content at Scale (https://contentatscale.ai/ai-content-detector/)
- Crossplag (https://crossplag.com/ai-content-detector/)
- DetectGPT (https://detectgpt.ericmitchell.ai/)
- Go Winston (https://gowinston.ai)
- GPT Zero (https://gptzero.me/)
- GPT-2 Output Detector Demo (https://openai-openai-detector.hf.space/)
- OpenAI Text Classifier (https://platform.openai.com/ai-text-classifier)
- PlagiarismCheck (https://plagiarismcheck.org/)
- Turnitin (https://demo-ai-writing-10.turnitin.com/home/)
- Writeful GPT Detector (https://x.writefull.com/gpt-detector)
- Writer (https://writer.com/ai-content-detector/)
- ZeroGPT (https://www.zerogpt.com/)

Table 2 gives an overview of the minimum/maximum sizes of text that could be examined by the free tools at the time of testing, if known.

Table 2. Minimum and maximum sizes for free tools

| Tool name | Minimum Size | Maximum Size |
|---|---|---|
| Check for AI | 350 characters | 2500 characters |
| Compilatio | 200 characters | 2000 characters |
| Content at Scale | 25 words | 25000 characters |
| Crossplag | Not stated | 1000 words |
| DetectGPT | 40 words | 256 words |
| Go Winston | 500 characters | 2000 words |
| GPT Zero | 250 characters | 5000 characters |
| GPT-2 Output Detector Demo | 50 tokens | 510 tokens |
| OpenAI Text Classifier | 1000 characters | Not stated |
| Writeful GPT Detector | 50 words | 1000 words |

| Writer | Not stated | 1500 characters |
| --- | --- | --- |
| ZeroGPT | Not stated | Not stated |

PlagiarismCheck and Turnitin are combined text similarity detectors and offer an additional functionality of determining the probability the text was written by an AI, so there was no limit on the amount of text tested. Signup was necessary for Check for AI, Crossplag, Go Winston, GPT Zero, and OpenAI Text Classifier (a Google account worked).

## 4.3 Data collection

A pilot study was undertaken on February 27, 2023, with one of the researchers submitting a fresh test document to each of the tools. The researcher shared their screen in Zoom with the other authors so that all could see how the tools worked. At this point, some tools were excluded because they required payment or because there was no answer from the tool. Some of the tools requested that the user rate the quality of the answer, but since the researchers did not want to provide additional training data to the tools, it was decided that no feedback would be given to any of the tools.

Since the authors were using Google Drive for sharing documents, all of the test cases were uploaded to folders for each of the categories and a Google form was prepared for data collection. In addition to selecting the tool, the category, the author of the test case, and the tester, there were free-text fields for noting down additional information such as how much of the text needed to be removed, what happened during the test, how long it took, and whether the result indicated that the author was a human or a bot.

On March 6, 2023, the authors gathered again in a Zoom room to test their own version of 01-Hum with Compilatio, Crossplag, and GPT Zero. This was to ensure that all testers were using the recording document in the correct manner and were putting their results in the correct fields. It was planned to record the time needed for each test, but since most of the results were returned quite quickly, it was decided to discontinue filling out this field. The timestamps for form submission were collected automatically.

The authors used their own test cases with each of the tools, recording the results in the standard form. The remaining test cases were split amongst the researchers for testing. If a tool only accepted fewer characters than the text to be tested, full sentences were removed from the end of the text until the text was less than the maximum amount. It was felt that preserving the integrity of paragraphs or at least sentences would be important for this type of testing.

The tests were run by the individual authors between March 7 and March 28, 2023. Since Turnitin was not available until April, those tests were completed between April 14 and April 20, 2023. The testing of PlagiarismCheck was performed between May 2 and May 8, 2023. Since Turnitin and PlagiarismCheck both also returned similarity scores, these were recorded but not used in the evaluation.

The raw data was copied into a fresh sheet and cleaned. Botched attempts and inadvertent duplicates were removed. The OpenAI Text Classifier, for example, would not work in the late afternoon/early evening in Europe, presumably because of the heavy load on the servers in the USA. Other duplicates occurred when test cases were tested twice by mistake. In two cases the results of the repeated test gave different results. One test case with Check for AI reported a low risk of AI on one test, and a high one five days later. ZeroGPT reported AI-generated on both tests but with different percentages. This was not systematically investigated, but in further work, it should be seen how widely the tools vary on the same material tested at a later time.

It was also ensured that all of the 54 test cases had been presented to each of the tools for a total of 756 tests.

## 4.4 Evaluation methodology

For the evaluation, the authors were split into groups of two or three and tasked with evaluating the results of the tests for the cases from either 01-Hum & 04-AI, 02-MT & 05-ManEd, or 03-AI & 06-Para. Since the tools do not provide an exact binary classification, one five-step classification was used for the original texts (01-Hum & 02-MT) and another one was used for the AI-generated texts (03-AI, 04-AI, 05-ManEd & 06-Para). They were based on the probabilities that were reported for texts being human-written or AI-generated as specified in Table 3.

Table 3. Classification accuracy scales for human-written and AI-generated texts

| **Human-written (NEGATIVE) text (docs 01-Hum & 02-MT), and the tool says that it is written by a:** | | |
|---|---|---|
| [100 - 80%) human | True negative | TN |
| [80 - 60%) human | Partially true negative | PTN |
| [60 - 40%) human | Unclear | UNC |
| [40 - 20%) human | Partially false positive | PFP |
| [20 - 0%] human | False positive | FP |
| **AI-generated (POSITIVE) text (docs 03-AI, 04-AI, 05-ManEd & 06-Para), and the tool says it is written by a:** | | |
| [100 - 80%) human | False negative | FN |
| [80 - 60%) human | Partially false negative | PFN |
| [60 - 40%) human | Unclear | UNC |
| [40 - 20%) human | Partially true positive | PTP |
| [20 - 0%] human | True positive | TP |

[ or ] means inclusive ( or ) means exclusive

For four of the detection tools, the results were only given in the textual form ("very low risk", "likely AI-generated", "very unlikely to be from GPT-2", etc.) and these were mapped to the classification labels as given in Table 4.

Table 4: Mapping of textual results to classification labels

| Tool | Result | 01-Hum, 02-MT | 03-AI, 04-AI, 05-ManEd, 06-Para |
|---|---|---|---|
| **Check for AI** | "very low risk" | TN | FN |
| | "low risk" | PTN | PFN |
| | "medium risk" | UNC | UNC |
| | "high risk" | PFP | PTP |
| | "very high risk" | FP | TP |
| **GPT Zero** | "likely to be written entirely by human" | TN | FN |
| | "may include parts written by AI" | PFP | PTP |
| | "likely to be written entirely by AI" | FP | TP |
| **OpenAI Text Classifier** | "The classifier considers the text to be …" | | |
| | "… likely AI-generated." | FP | TP |
| | "… possibly AI-generated." | PFP | PTP |
| | "Unclear if it is AI-generated" | UNC | UNC |
| | "… unlikely AI-generated." | PTN | PFN |
| | "… very unlikely AI-generated." | TN | FN |
| **DetectGPT** | "very unlikely to be from GPT-2" | TN | FN |
| | "unlikely to be from GPT-2" | PTN | PFN |
| | "likely to be from GPT-2" | PFP | PTP |
| | "very likely from GPT-2" | FP | TP |

After all of the classifications were undertaken and disagreements ironed out, the measures of accuracy, the false positive rate, and the false negative rate were calculated.

# 5. Results

Having evaluated the classification outcomes of the tools as (partially) true/false positives/negatives, the researchers evaluated this classification on two criteria: accuracy and error type. In general, classification systems are evaluated using accuracy, precision, and recall. The research authors also conducted an error analysis since the educational context means different types of error have different significance.

## 5.1 Accuracy

When no partial results are allowed, i.e. only TN, TP, FN, and FP are allowed, accuracy is defined as a ratio of correctly classified cases to all cases

$$ACC = (TN + TP) / (TN + TP + FN + FP);$$

There are several ways of dealing with partially (in)correct classifications when calculating accuracy. As the denominator of the formula contains ALL cases, the first approach is to consider partially correct classification as incorrect and calculate the accuracy as

$$ACC_bin = (TN + TP) / (TN + PTN + TP + PTP + FN + PFN + FP + PFP + UNC)$$

Table 5 shows the number of correctly classified documents, i.e. the sum of true positives and true negatives. The maximum for each cell is 9 (because there were 9 documents in each class), the overall maximum is 9 × 6 = 54. The accuracy is calculated as a ratio of the total and the overall maximum. Note that even the highest accuracy values are below 80%. The last row shows the average accuracy for each document class, across all the tools.

Table 5. Accuracy of the detection tools (binary approach)

| Tool | 01-Hum | 02-MT | 03-AI | 04-AI | 05-ManEd | 06-Para | Total | Accuracy | Rank |
|---|---|---|---|---|---|---|---|---|---|
| Check For AI | 9 | 0 | 9 | 8 | 4 | 2 | 32 | 59% | 6 |
| Compilatio | 8 | 9 | 8 | 8 | 5 | 2 | 40 | 74% | 2 |
| Content at Scale | 9 | 9 | 0 | 0 | 0 | 0 | 18 | 33% | 14 |
| Crossplag | 9 | 6 | 9 | 7 | 4 | 2 | 37 | 69% | 4 |
| DetectGPT | 9 | 5 | 2 | 8 | 0 | 1 | 25 | 46% | 11 |
| Go Winston | 7 | 7 | 9 | 8 | 4 | 1 | 36 | 67% | 5 |
| GPT Zero | 6 | 3 | 7 | 7 | 3 | 3 | 29 | 54% | 8 |
| GPT-2 Output Detector Demo | 9 | 7 | 9 | 8 | 5 | 1 | 39 | 72% | 3 |

| OpenAI Text Classifier | 9 | 8 | 2 | 7 | 2 | 1 | 29 | 54% | 8 |
|---|---|---|---|---|---|---|---|---|---|
| PlagiarismCheck | 7 | 5 | 3 | 3 | 1 | 2 | 21 | 39% | 13 |
| Turnitin | 9 | 9 | 8 | 9 | 4 | 2 | 41 | 76% | 1 |
| Writeful GPT Detector | 9 | 7 | 2 | 3 | 2 | 0 | 23 | 43% | 12 |
| Writer | 9 | 7 | 4 | 4 | 2 | 1 | 27 | 50% | 10 |
| ZeroGPT | 9 | 5 | 7 | 8 | 2 | 1 | 32 | 59% | 6 |
| **Average** | **94%** | **69%** | **63%** | **70%** | **30%** | **15%** | | | |

This method provides a good overview of the number of cases in which the classifiers are "sure" about the outcome. However, for real-life educational scenarios, partially correct classifications are also valuable. Especially in case 05-ManEd, which involved human editing, the partially positive classification results make sense. Therefore, the researchers explored two more ways of assessment. These methods differ in the score awarded to various incorrect outcomes.

In semi-binary evaluation, partially correct classification (PTN or PTP) was awarded 0.5 points, while entirely correct classification (TN or TP) gained 1.0 points as in the previous method. The formula for accuracy calculation is

$$ACC_SEMIBIN = (TN + TP + 0.5 \times PTN + 0.5 \times PTP) / (TN + PTN + TP + PTP + FN + PFN + FP + PFP + UNC)$$

Table 6 shows the assessment results of the classifiers using semi-binary classification. The values correspond to the number of correctly classified documents with partially correct results awarded half a point (TP + TN + 0.5 × PTN + 0.5 × PTP). The maximum value is again 9 for each cell and 54 for the total.

Table 6. Accuracy of the detection tools (semi-binary approach)

| Tool | 01-Hum | 02-MT | 03-AI | 04-AI | 05-ManEd | 06-Para | Total | Accu-racy | Rank |
|---|---|---|---|---|---|---|---|---|---|
| Check For AI | 9 | 3.5 | 9 | 8 | 4 | 2.5 | 36 | 67% | 6 |
| Compilatio | 8.5 | 9 | 8.5 | 8 | 5.5 | 2 | 41.5 | 77% | 2 |
| Content at Scale | 9 | 9 | 0 | 0 | 0 | 0 | 18 | 33% | 14 |
| Crossplag | 9 | 6 | 9 | 7 | 4.5 | 2 | 37.5 | 69% | 5 |
| DetectGPT | 9 | 6.5 | 5.5 | 8 | 2 | 1.5 | 32.5 | 60% | 10 |
| Go Winston | 7.5 | 7.5 | 9 | 8 | 4.5 | 1.5 | 38 | 70% | 4 |
| GPT Zero | 6 | 3 | 7.5 | 8 | 5.5 | 5.5 | 35.5 | 66% | 8 |
| GPT-2 Output Detector Demo | 9 | 7 | 9 | 8 | 5 | 1.5 | 39.5 | 73% | 3 |
| OpenAI Text Classifier | 9 | 8.5 | 3.5 | 7.5 | 3.5 | 1.5 | 33.5 | 62% | 9 |
| PlagiarismCheck | 8 | 6.5 | 4 | 4.5 | 2 | 2.5 | 27.5 | 51% | 13 |
| Turnitin | 9 | 9 | 8.5 | 9 | 4.5 | 2.5 | 42.5 | 79% | 1 |

| Writeful GPT Detector | 9 | 7.5 | 5 | 4.5 | 2.5 | 0.5 | 29 | 54% | 12 |
|---|---|---|---|---|---|---|---|---|---|
| Writer | 9 | 7 | 4.5 | 5 | 3 | 1.5 | 30 | 56% | 11 |
| ZeroGPT | 9 | 6.5 | 7 | 8 | 3 | 2.5 | 36 | 67% | 6 |
| **Average** | **95%** | **77%** | **71%** | **74%** | **39%** | **22%** | | | |

A semi-binary approach to accuracy calculation captures the notion of partially correct classification but still does not distinguish between various forms of incorrect classification. The researchers address this issue in a logarithmic approach to accuracy calculation that awards 1 point to completely incorrect classification and doubles the score for each level of the classification that was closer to the correct result. The scores for the particular classifier outputs are shown in Table 7 and the overall scores of the classifiers are shown in Table 8. Note that the maximum value for each cell is now 9 × 16 = 864. The accuracy, again, is calculated as a ratio of the total score and the maximum possible score. This approach provides the most detailed distinction among all varieties of (in)correctness.

Table 7. Scores for logarithmic evaluation

| Positive case | Negative case | Score |
|---|---|---|
| FN | FP | 1 |
| PFN | PFP | 2 |
| UNC | UNC | 4 |
| PTP | PTN | 8 |
| TP | TN | 16 |

Table 8. Logarithmic approach to accuracy evaluation

| Tool | 01-Hum | 02-MT | 03-AI | 04-AI | 05-ManEd | 06-Para | Total | Accuracy | Rank |
|---|---|---|---|---|---|---|---|---|---|
| Check For AI | 144 | 62 | 144 | 129 | 74 | 54 | 607 | 70% | 7 |
| Compilatio | 136 | 144 | 136 | 132 | 91 | 40 | 679 | 79% | 2 |
| Content at Scale | 144 | 144 | 23 | 24 | 17 | 18 | 370 | 43% | 14 |
| Crossplag | 144 | 99 | 144 | 115 | 76 | 40 | 618 | 72% | 6 |
| DetectGPT | 144 | 108 | 88 | 129 | 38 | 36 | 543 | 63% | 10 |
| Go Winston | 124 | 124 | 144 | 130 | 79 | 45 | 646 | 75% | 4 |
| GPT Zero | 102 | 60 | 121 | 128 | 89 | 89 | 589 | 68% | 8 |
| GPT-2 Output Detector Demo | 144 | 114 | 144 | 129 | 84 | 35 | 650 | 75% | 3 |
| OpenAI Text Classifier | 144 | 136 | 67 | 124 | 67 | 48 | 586 | 68% | 9 |
| PlagiarismCheck | 128 | 108 | 76 | 82 | 50 | 53 | 497 | 58% | 12 |
| Turnitin | 144 | 144 | 136 | 144 | 81 | 53 | 702 | 81% | 1 |

| Writeful GPT Detector | 144 | 122 | 81 | 76 | 50 | 20 | 493 | 57% | 13 |
|---|---|---|---|---|---|---|---|---|---|
| Writer | 144 | 117 | 83 | 84 | 53 | 35 | 516 | 60% | 11 |
| ZeroGPT | 144 | 108 | 120 | 132 | 65 | 54 | 623 | 72% | 5 |
| **Average** | **96%** | **79%** | **75%** | **77%** | **45%** | **31%** | | | |

As can be seen from Tables 5, 6 and 8, the approach to accuracy evaluation has almost no influence on the ranking of the classifiers. Figure 1 presents the overall accuracy for each tool as the mean of all accuracy approaches used.

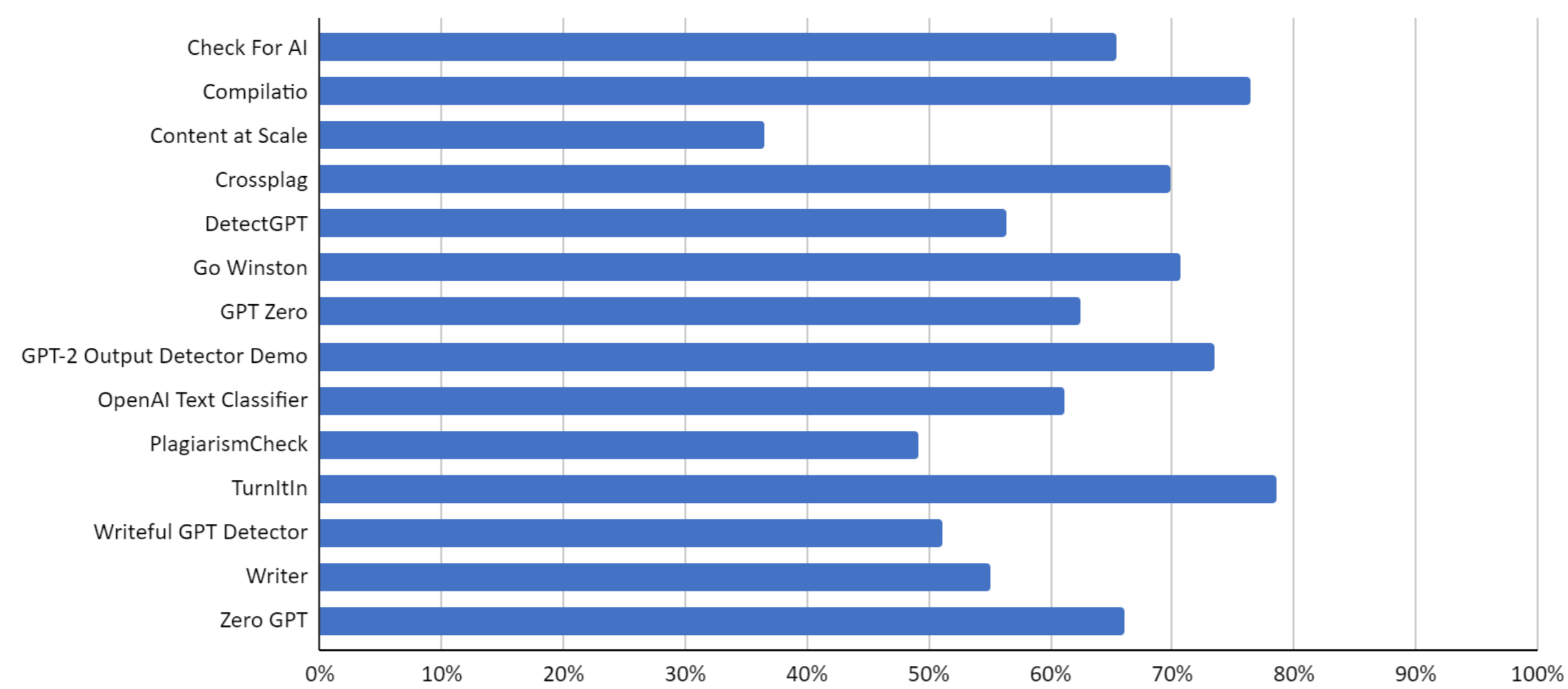

Figure 1: Overall accuracy for each tool calculated as an average of all approaches discussed

Turnitin received the highest score using all approaches to accuracy classification, followed by Compilatio and GPT-2 Output Detector (again in all approaches). This is particularly interesting because as the name suggests, GPT-2 Output Detector was not trained to detect GPT-3.5 output. Crossplag and Go Winston were the only other tools to achieve at least 70% accuracy.

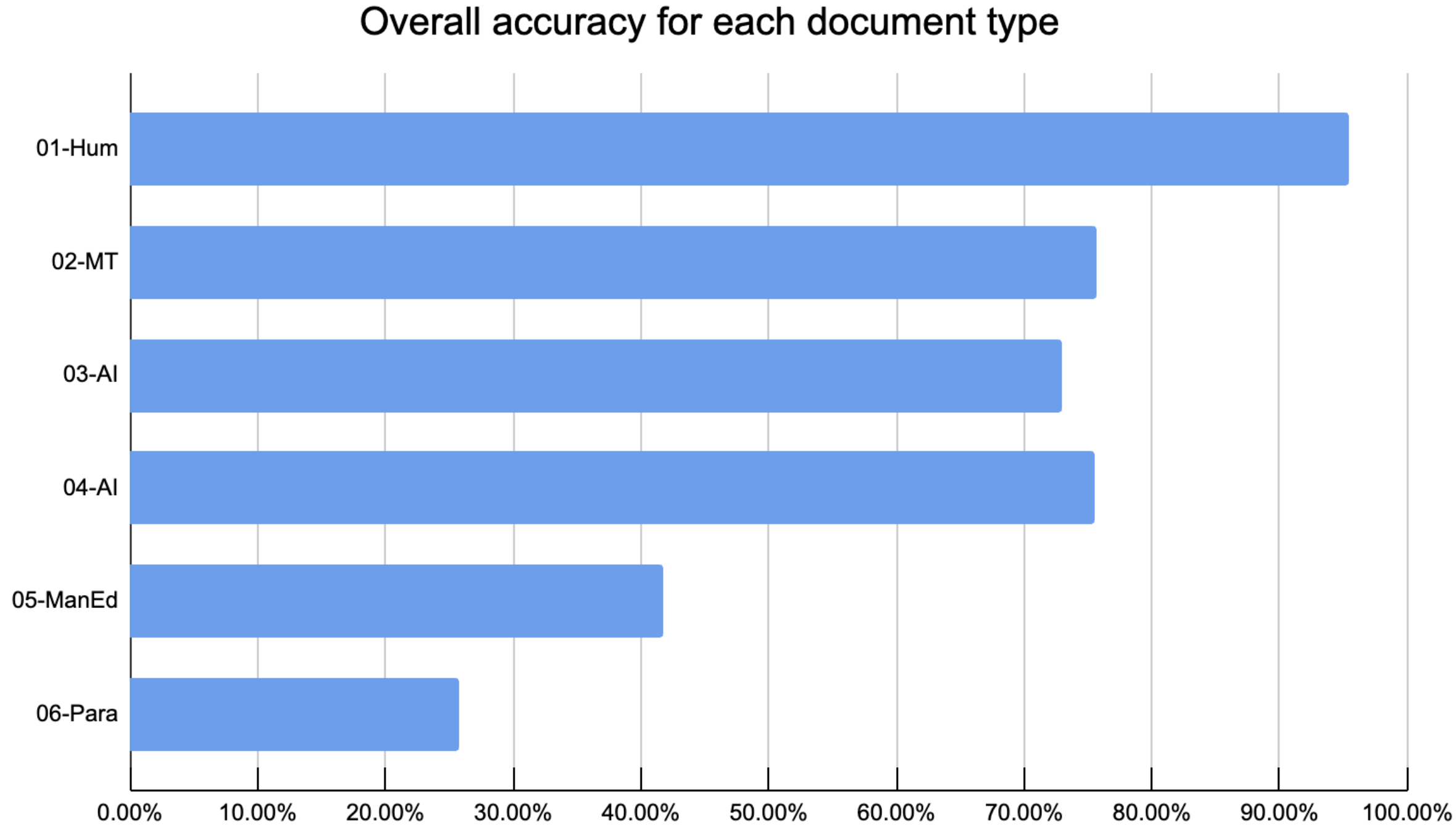

Figure 2: Overall accuracy for each document type (calculated as an average of all approaches discussed)

### 5.1.1 Variations in Accuracy

As Figure 2 above shows, the overall average accuracy figure is misleading, as it obscures major variations in accuracy between document types. Further analysis reveals the influence of machine translation, human editing, and machine paraphrasing on overall accuracy:

**Influence of machine translation:** The overall accuracy for case 01-Hum (human-written) was 96%. However, in the case of the documents written by humans in languages other than English that were machine-translated to English (case 02-MT), the accuracy dropped by 20%. Apparently, machine translation leaves some traces of AI in the output, even if the original was purely human-written.

**Influence of human manual editing**: Case 05-ManEd (machine-generated with subsequent human editing) generally received slightly over half the score (42%) compared to cases 03-AI and 04-AI (machine-generated with no further modifications; 74%). This reflects a typical scenario of student misconduct in cases where the use of AI is prohibited. The student obtains a text written by an AI and then quickly goes through it and makes some minor changes such as using synonyms to try to disguise unauthorised content generation. This type of writing has been called patchwriting (Howard, 1995). Only ~50% accuracy of the classifiers shows that these cases, which are assumed to be the most common ones, are almost undetectable by current tools.

**Influence of machine paraphrase:** Probably the most surprising results are for case 06-Para (machine-generated with subsequent machine paraphrase). The use of AI to transform AI-generated text results in text that the classifiers consider human-written. The overall accuracy for this case was 26%, which means that most AI-generated texts remain undetected when machine-paraphrased.

### 5.1.2 Consistency in tool results

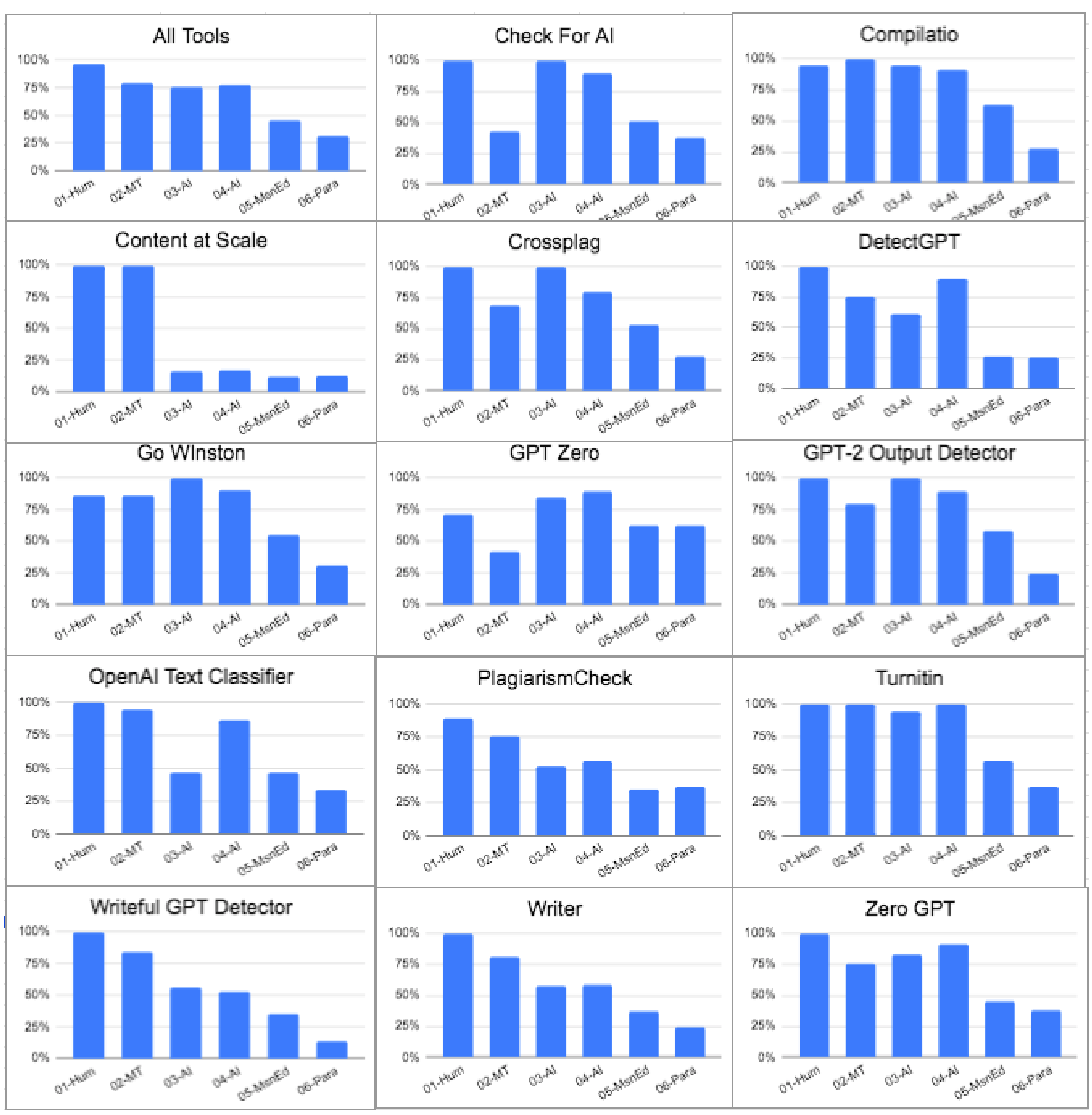

Figure 3: Accuracy (logarithmic) for each document type by detection tool for AI-generated text.

With the notable exception of GPT Zero, all the tested tools followed the pattern of higher accuracy when identifying human-written text than when identifying texts generated or modified by AI or machine tools, as seen in Figure 3. Therefore, their classification is (probably deliberately) biased towards humans rather than AI output. This classification bias is preferable in academic contexts for the reasons discussed below.

## 5.2 Error analysis

Analysis of classification errors is an integral part of this research. The researchers focused on two types of errors that might have significant consequences in educational contexts: false accusations against a student and a student gaining an unfair advantage over others.

### 5.2.1 False Accusations: Harm to individual students

If educators use one of the classifiers to detect student misconduct, there is a question of what kind of output leads to the accusation of a student from unauthorised content generation. The researchers believe that a typical educator would accuse a student if the output of the classifier is positive or partially positive. Some teachers may also suspect students of misconduct in unclear or partially negative cases, but the research authors think that educators generally do not initiate disciplinary action in these cases. Therefore, for each tool, we also computed the likelihood of false accusation of a student as a ratio of false positives and partially false positives to all cases, i.e.

$$FAS = (FP + PFP) / (TN + PTN + TP + PTP + FN + PFN + FP + PFP + UNC)$$

Table 9. False accusation ratio

| Tool | 01-Hum | 02-MT | Total | FAS ratio |
|---|---|---|---|---|
| Check For AI | 0 | 1 | 1 | 5.6% |
| Compilatio | 0 | 0 | 0 | 0.0% |
| Content at Scale | 0 | 0 | 0 | 0.0% |
| Crossplag | 0 | 3 | 3 | 16.7% |
| DetectGPT | 0 | 0 | 0 | 0.0% |
| Go Winston | 0 | 0 | 0 | 0.0% |
| GPT Zero | 3 | 6 | 9 | 50.0% |
| GPT-2 Output Detector Demo | 0 | 2 | 2 | 11.1% |
| OpenAI Text Classifier | 0 | 0 | 0 | 0.0% |
| PlagiarismCheck | 0 | 0 | 0 | 0.0% |
| Turnitin | 0 | 0 | 0 | 0.0% |
| Writeful GPT Detector | 0 | 1 | 1 | 5.6% |
| Writer | 0 | 1 | 1 | 5.6% |
| ZeroGPT | 0 | 0 | 0 | 0.0% |
| **Average** | **2.4%** | **11.1%** | | |

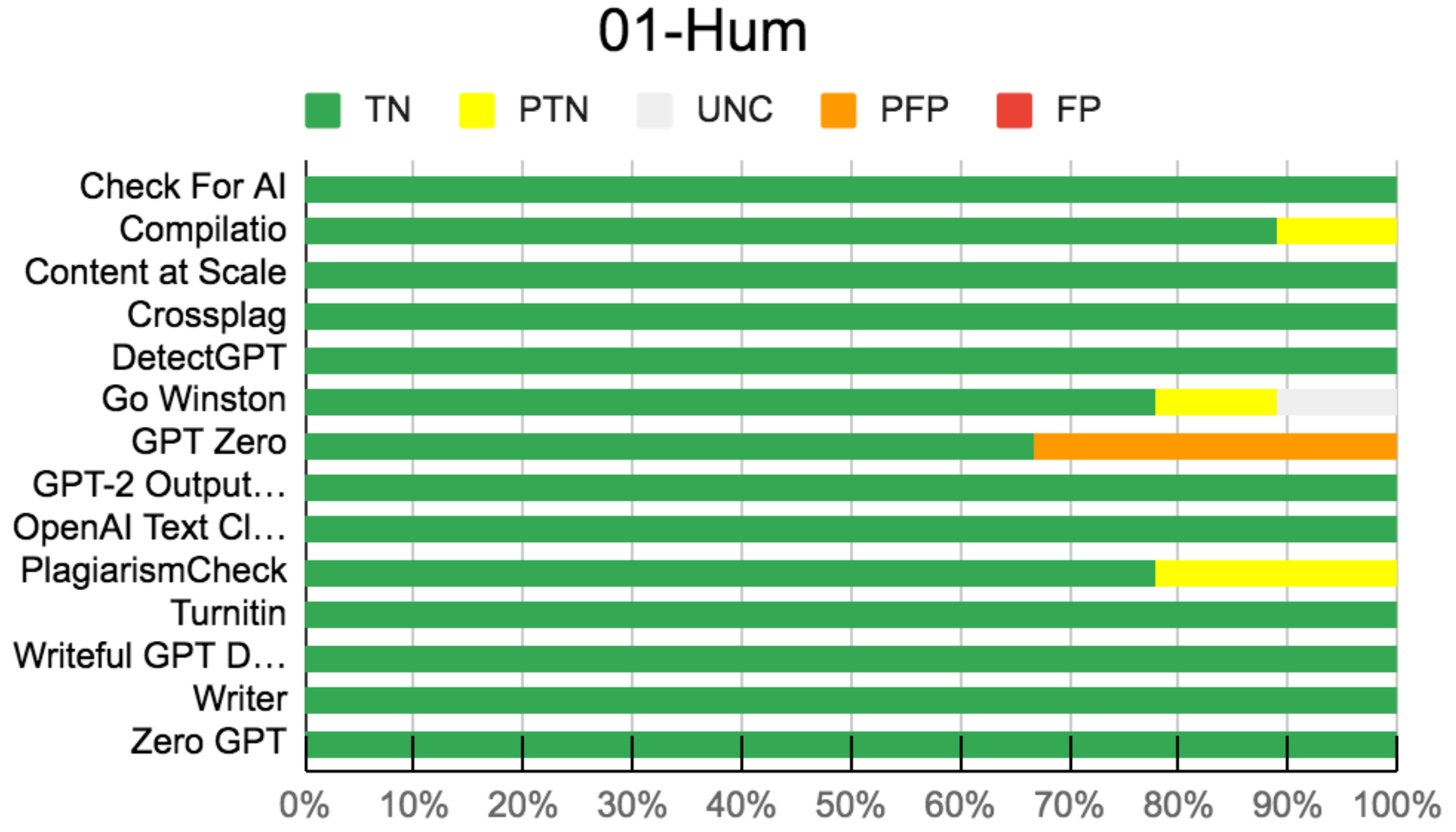

Figure 4: False accusations for human-written documents

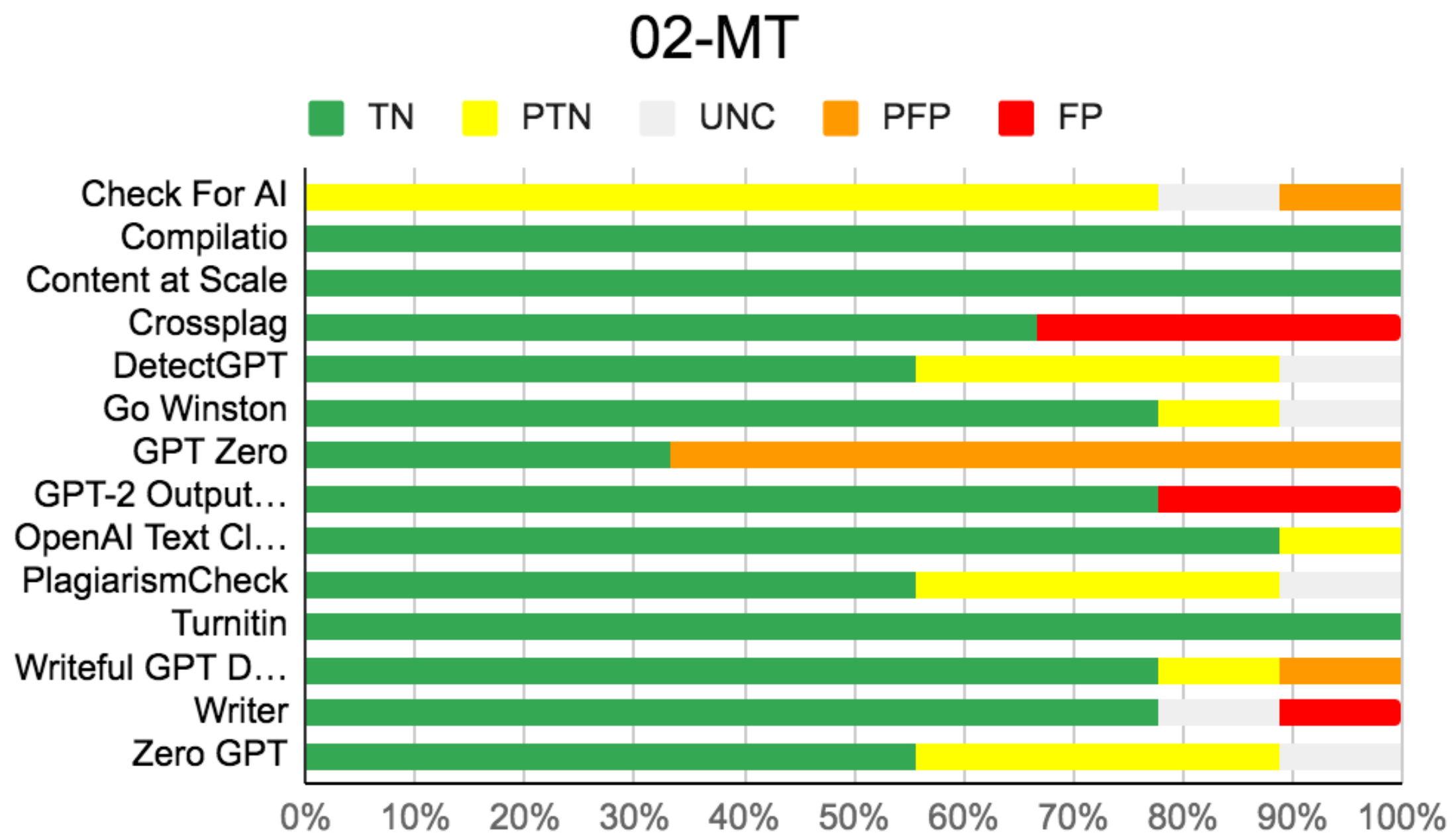

Figure 5: False accusations for machine-translated documents

Table 9 shows the number of cases in which the classification of a particular document would lead to a false accusation. The table includes only documents 01-Hum and 02-MT, because the AI-generated documents are not relevant. The risk of false accusations is zero for half of the tools, as can be also seen from Figures 4 and 5. Six of the fourteen tools tested generated false positives, with the risk increasing dramatically for machine-translated texts. For GPT Zero, half of the positive classifications would be false accusations, which makes this tool unsuitable for the academic environment.

## 5.2.2 Undetected cases: Undermining academic integrity

Another form of academic harm is undetected cases, i.e. AI-generated texts that remain undetected. A student who used unauthorised content generation likely obtains an unfair advantage over those who fulfilled the task with integrity. The actual victims of this form of misconduct are the honest students that receive the same credits as the dishonest ones. The likelihood of an AI-generated document being undetected is given in Table 10, which includes only cases 03-AI, 04-AI, 05-ManEd and 06-Para.

Table 10. Percentage of undetected cases

| Tool | 03-AI | 04-AI | 05-ManEd | 06-Para | Total | UDC ratio |
|---|---|---|---|---|---|---|
| Check For AI | 0 | 1 | 5 | 6 | 12 | 33.3% |
| Compilatio | 0 | 1 | 3 | 7 | 11 | 30.6% |
| Content at Scale | 9 | 9 | 9 | 9 | 36 | 100.0% |
| Crossplag | 0 | 2 | 4 | 7 | 13 | 36.1% |
| DetectGPT | 0 | 1 | 5 | 7 | 13 | 36.1% |
| Go Winston | 0 | 1 | 4 | 7 | 12 | 33.3% |
| GPT Zero | 1 | 0 | 1 | 1 | 3 | 8.3% |
| GPT-2 Output Detector Demo | 0 | 1 | 4 | 7 | 12 | 33.3% |
| OpenAI Text Classifier | 4 | 1 | 4 | 7 | 16 | 44.4% |
| PlagiarismCheck | 4 | 3 | 6 | 6 | 19 | 52.8% |
| Turnitin | 0 | 0 | 4 | 6 | 10 | 27.8% |
| Writeful GPT Detector | 1 | 3 | 6 | 8 | 18 | 50.0% |
| Writer | 4 | 3 | 5 | 7 | 19 | 52.8% |
| ZeroGPT | 2 | 1 | 5 | 5 | 13 | 36.1% |
| **Average** | **19.8%** | **21.4%** | **51.6%** | **71.4%** | | |

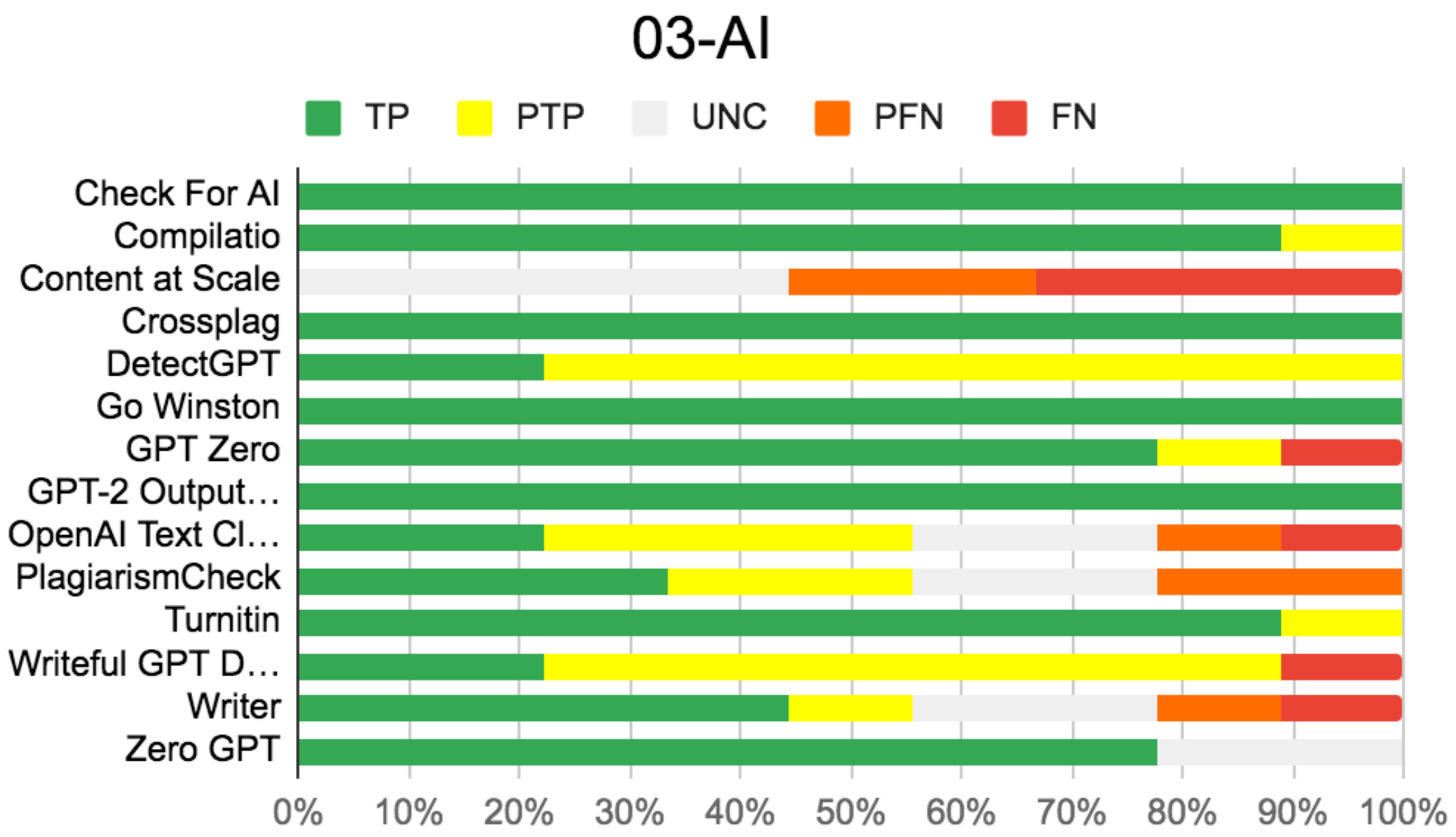

Figure 6: False negatives for AI-generated documents 03-AI

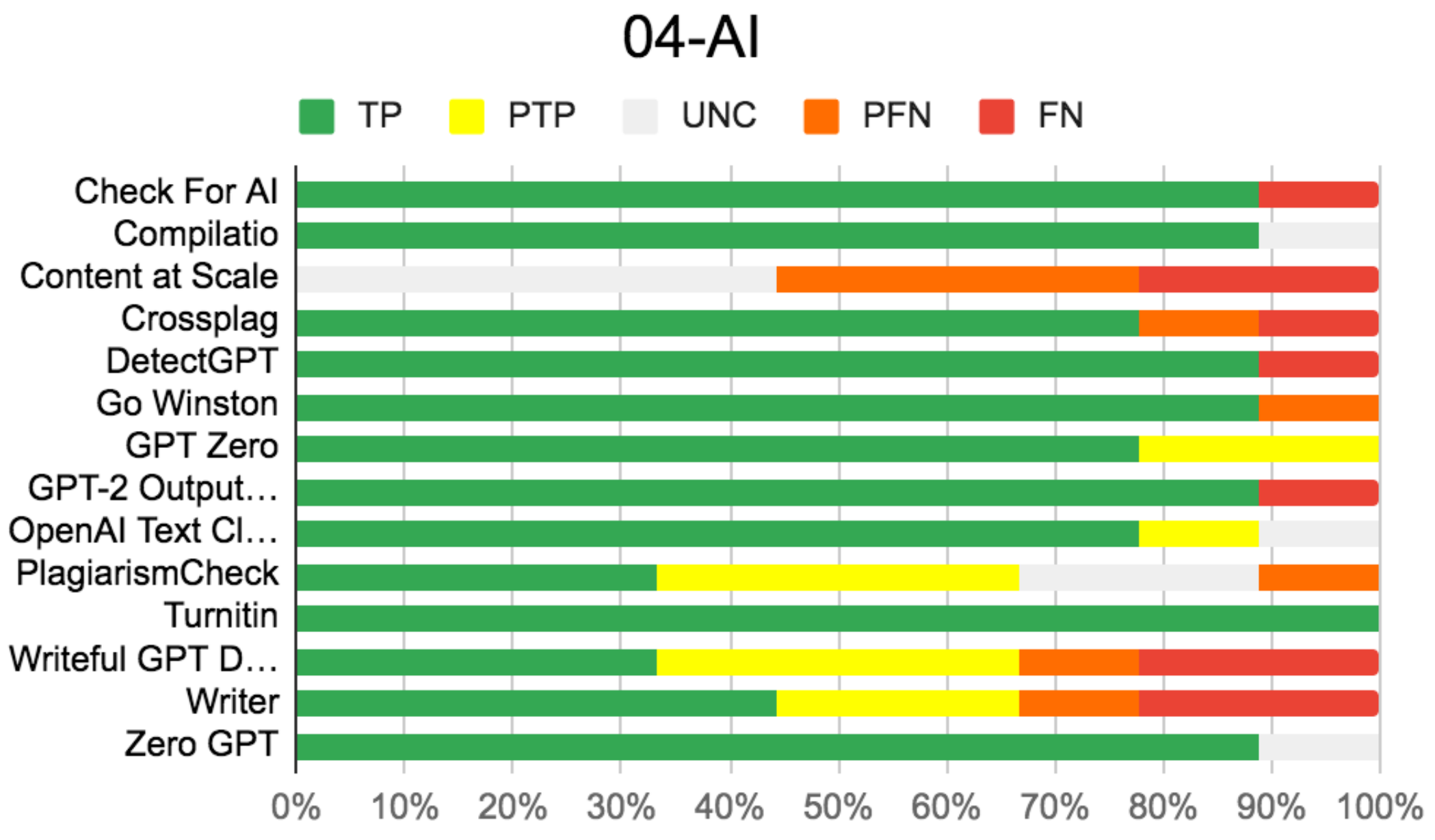

Figure 7: False negatives for AI-generated documents 04-AI

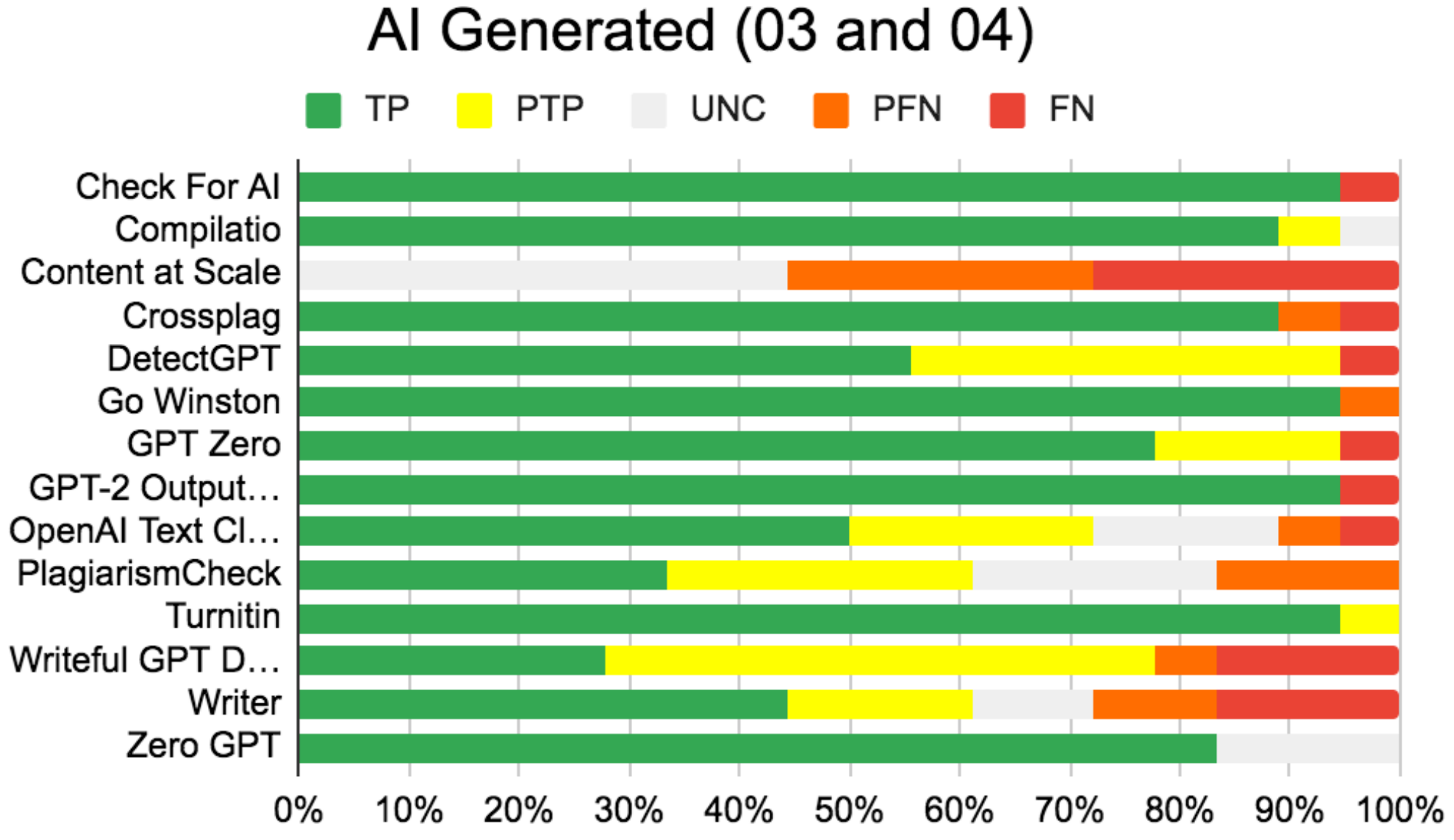

Figure 8: False negatives for AI-generated documents 03-AI and 04-AI together

Figures 6, 7, and 8 above show that 13 out of the 14 tested tools produced false negatives or partially false negatives for documents 03-AI and 04-AI; only Turnitin correctly classified all documents in these classes. None of the tools could correctly classify all AI-generated documents that undergo manual editing or machine paraphrasing.

As the document sets 03-AI and 04-AI were prepared using the same method, the researchers expected the results would be the same. However, for some tools (OpenAI Text Classifier and DetectGPT), the results were notably different. This could indicate a mistake in testing made or interpretation of the results. Therefore, the researchers double-checked all the results to avoid this kind of mistake. We also tried to upload some documents again. We did obtain different values, but we found out that this was due to inconsistency in the results of these tools and not due to our mistakes.

Content at Scale misclassified all of the positive cases; these results in combination with the 100% correct classification of human-written documents indicate that the tool is inherently biased towards human classification and thus completely useless. Overall, of the AI-generated texts approx. 20% of cases would likely be misattributed to humans, meaning the risk of unfair advantage is significantly greater than that of false accusation.

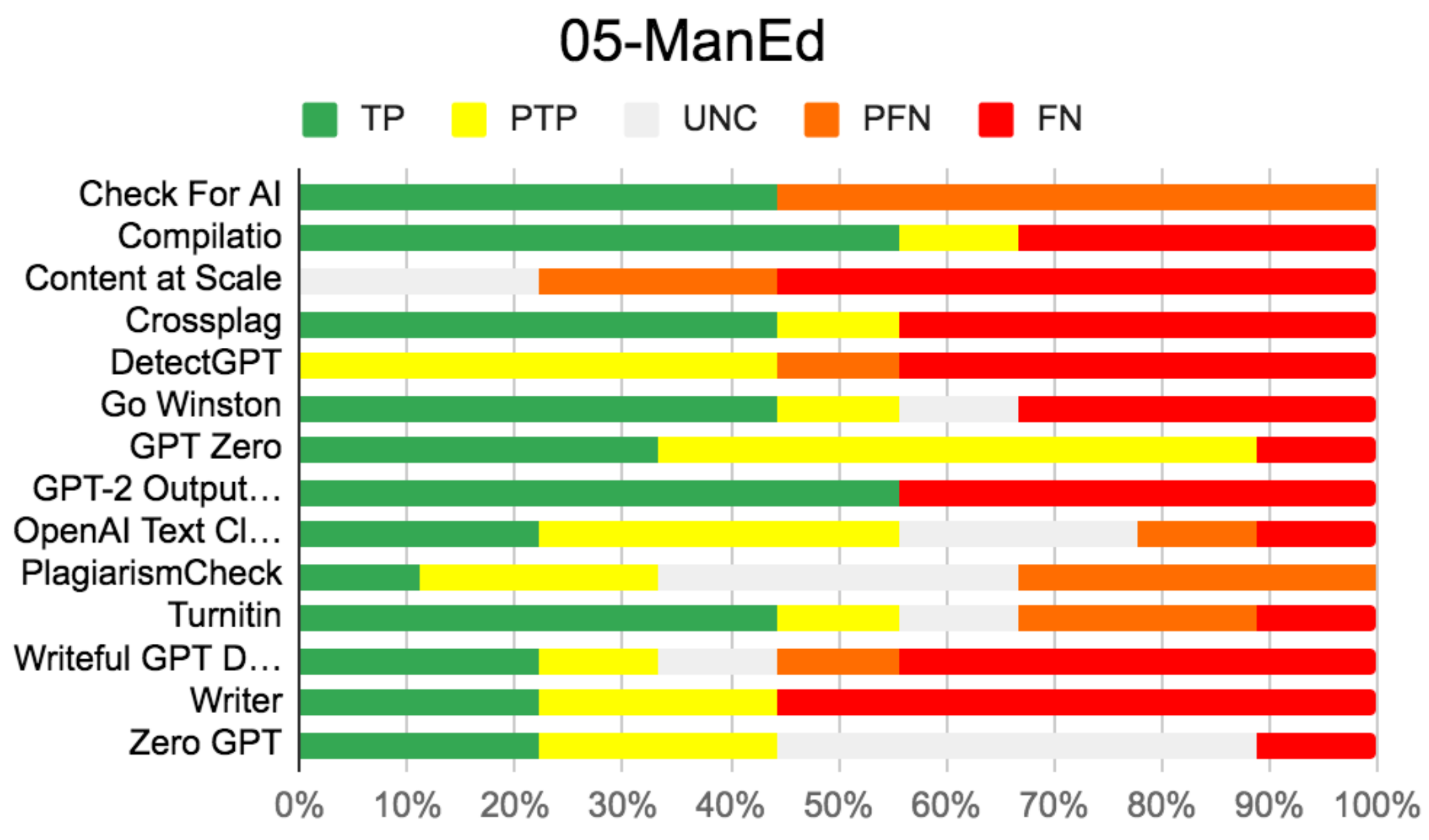

Figure 9: False negatives for manually edited documents

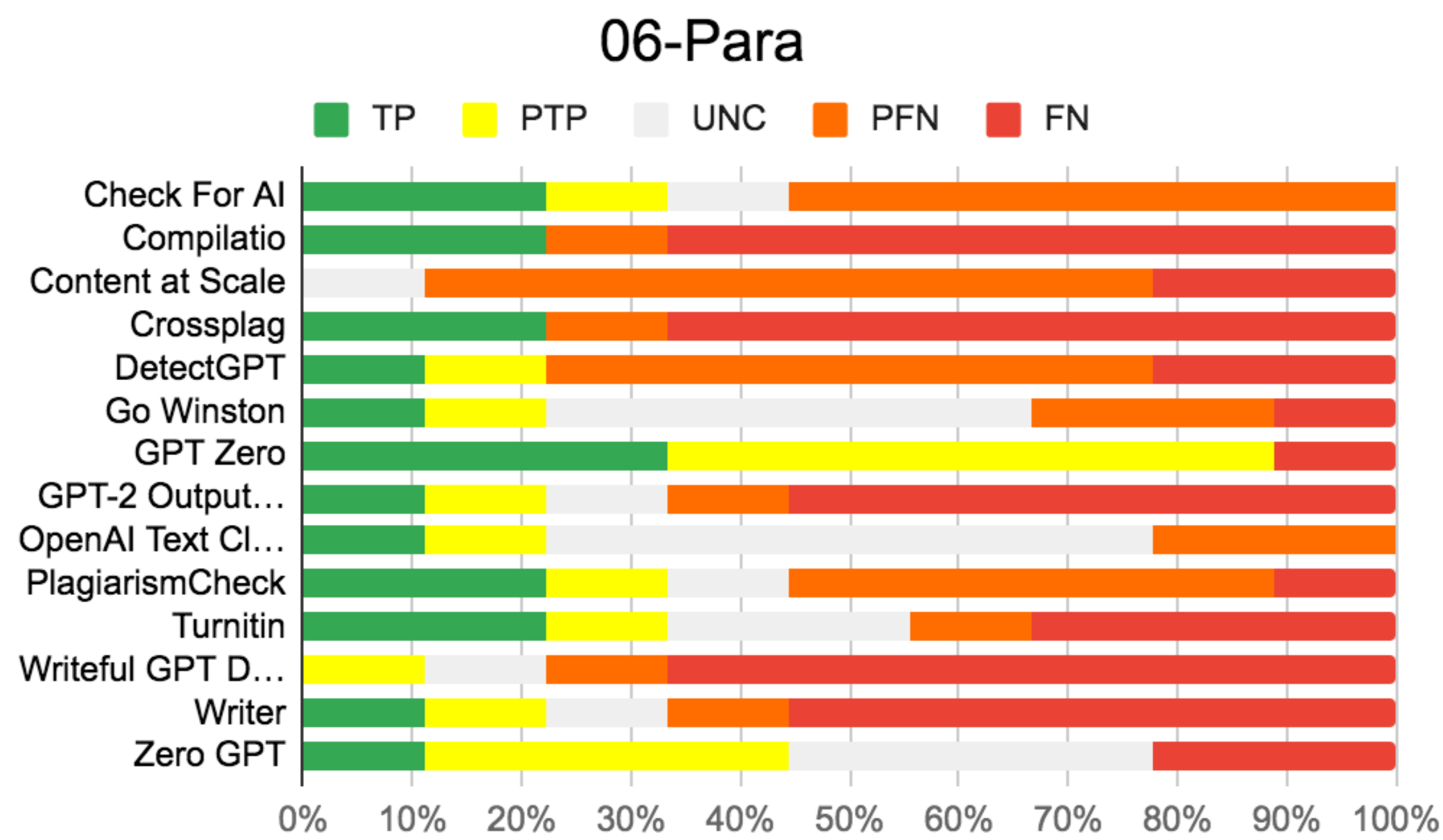

Figure 10: False negatives for machine-paraphrased documents

Figures 9 and 10 show an even greater risk of students gaining an unfair advantage through the use of obfuscation strategies. At an overall level, for manually edited texts (case 05-ManEd) the ratio of undetected texts increases to approx. 50% and in the case of machine-paraphrased texts (case 06-Para) rises even higher.

## 5.3 Usability issues

There were a few usability issues that cropped up during the testing that may be attributable to the beta nature of the tools under investigation.

For example, the tool DetectGPT at some point stopped working and only replied with the statement "Server error 😕 We might just be overloaded. Try again in a few minutes?". This issue occurred after the initial testing round and persisted until the time of submission of this paper. Others would stall in an apparent infinite loop or throw an error message and the test had to be repeated at a later time.

Writeful GPT Detector would not accept computer code. The tool apparently identified code as not English, and the tool only accepted English texts.

Compilatio at one point returned "NaN% reliability" (See Figure 11) for a ChatGPT-generated text that included program code. "NaN" is computer jargon for "not a number" and indicates that there were calculation issues such as division by zero or number representation overflow. Since there was also a robot head returned, this was evaluated as correctly identifying ChatGPT-generated text, but the non-numerical percentage might confuse instructors using the tool.

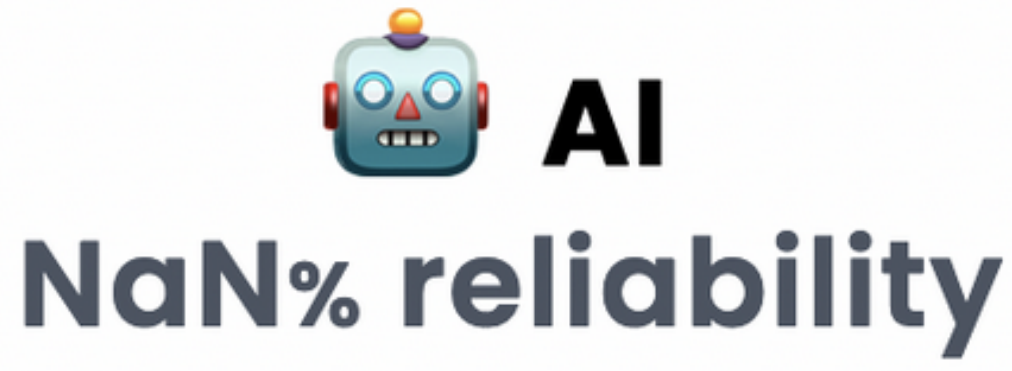

Figure 11: Compilatio's NaN% reliability

The operation of a few of the tools was not immediately clear to some of the authors and the handling of results was sometimes not easy to document. For example, in PlagiarismCheck the AI-Detection button was not always presented on the screen and it would only show the last four tests done. Interestingly, Turnitin often returned high similarity values for ChatGPT-generated text, especially for program code or program output. This was distracting, as the similarity results were given first, the AI-detection could only be accessed by clicking on a number above the text "AI" that did not look clickable, but was, see Figure 12.

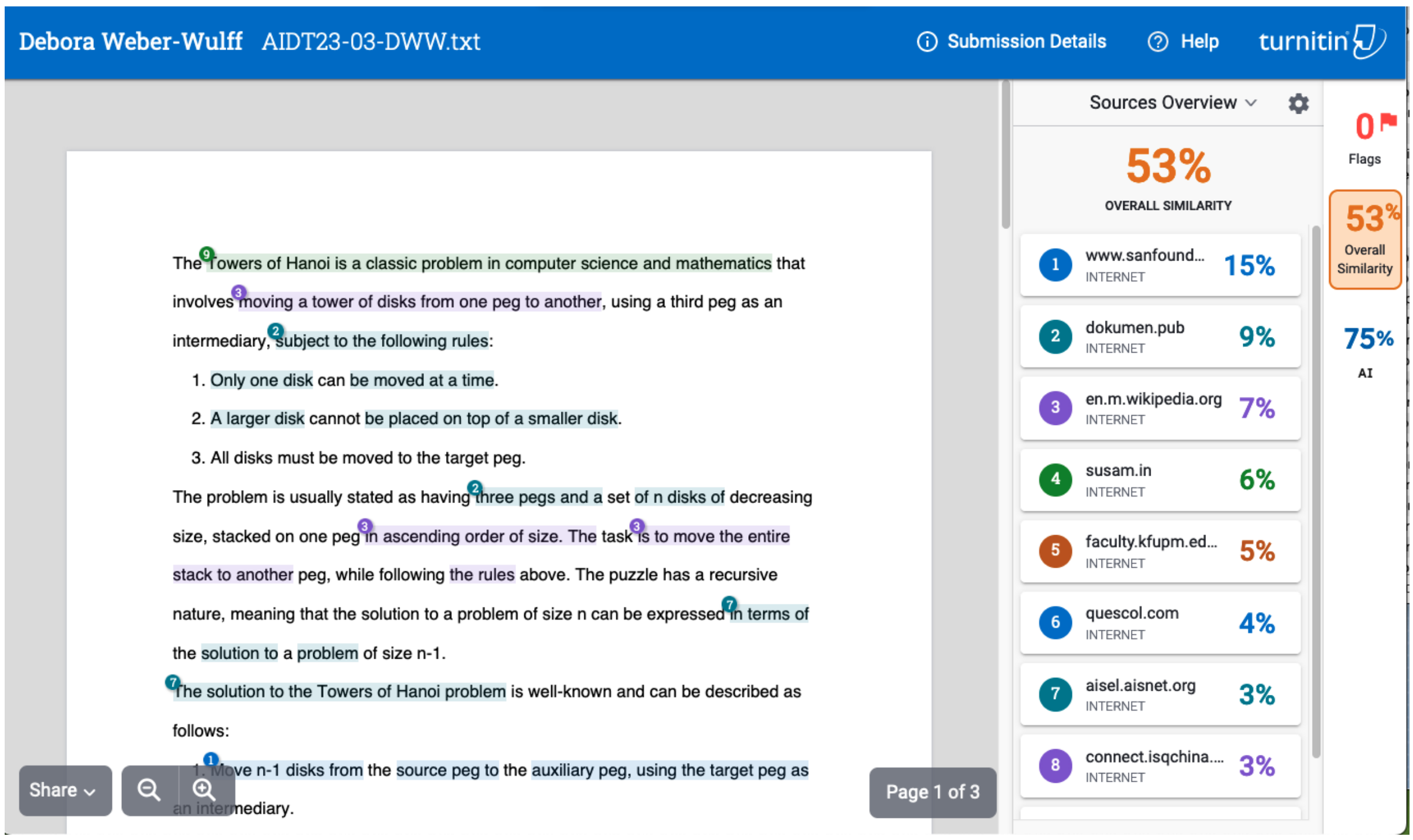

Figure 12: Turnitin's similarity report shows up first, it is not clear that the "AI" is clickable

# 6. Discussion

Detection tools for AI-generated text do fail, they are neither accurate nor reliable (all scored below 80% of accuracy and only 5 over 70%). In general, they have been found to diagnose human-written documents as AI-generated (false positives) and often diagnose AI-generated texts as human-written (false negatives). Our findings are consistent with previously published studies (Gao et al., 2022; Anderson et al., 2023; Demers, 2023; Gewirtz, 2023; Krishna et al., 2023; Pegoraro et al., 2023; van Oijen, 2023; Wang et al., 2023) and substantially differ from what some detection tools for AI-generated text claim (Compilatio, 2023; Crossplag.com, 2023; GoWinston.ai, 2023; ZeroGPT, 2023). The detection tools present a main bias towards classifying the output as human-written rather than detecting AI-generated content. Overall, approximately 20% of AI-generated texts would likely be misattributed to humans.

They are neither robust, since their performance worsens even more with the use of obfuscation techniques such as manual editing or machine paraphrasing, nor are they able to cope with texts translated from other languages. Overall, approximately 50% of AI-generated texts that undergo some obfuscation would likely be misattributed to humans.

The results provided by the tools are not always easy to interpret for an average user. Some of them provide statistical information to justify the classification, and others highlight the text that is "likely" machine-generated. Some present values such as "perplexity = 137.222" or "Burstiness Score: 17104.959" with many digits of precision that do not generally help a user understand the results.

Some of the detection tools such as Writer are clearly aimed to be used to hide AI-written text, providing suggestions to users such as “You should edit your text until there’s less detectable AI content.” (See Figure 13)

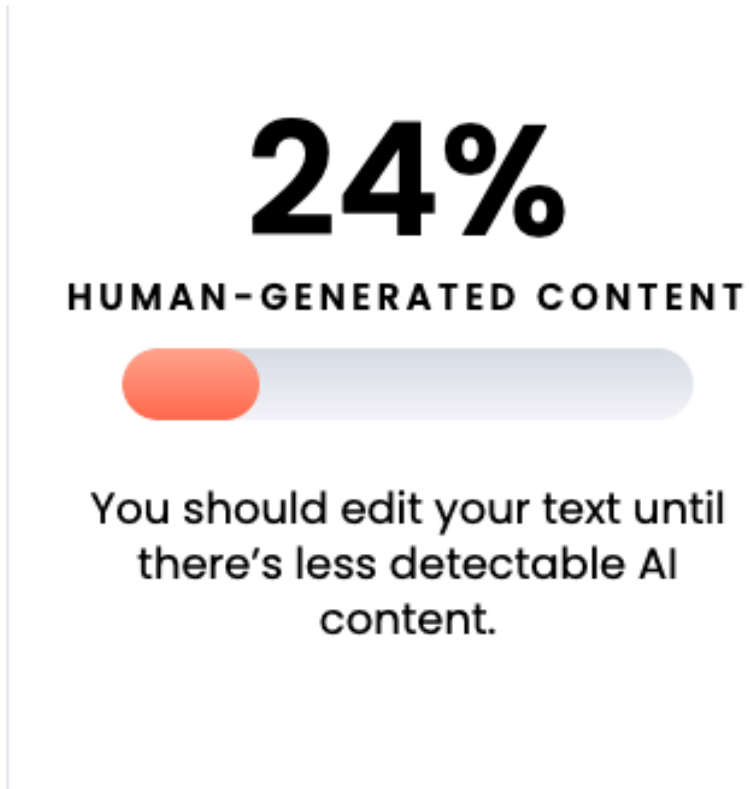

Figure 13: Writer’s suggestion to lower “detectable AI content”

Detection tools for AI-generated text provide simple outputs with statements like “This document was likely written by AI” or “11% likely this comes from GPT-3, GPT-4 or ChatGPT”, without any possibility of verification or evidence. Therefore, a student accused of unauthorised content generation only on this basis would have no possibility for a defence. The probability of false positives ranged from 0% (Turnitin) to 50% (GPT Zero). The probability of false negatives ranged from 8% (GPT Zero) to 100% (Content at Scale). The different types of failures may have serious implications. False positives could lead to wrong accusations of students, the false negatives allow students to evade detection of unauthorised content generation gaining unfair advantages and promoting impunity. Our experience and personal communications indicate that there is a large group of academics that believe in the output of the classifiers. The research results show that users should be extremely cautious when interpreting the results.

It is noteworthy that using machine translation such as Google translate or DeepL can lead to a higher number of false positives, leaving L2 students (and researchers) at risk of being falsely accused of unauthorised content generation when using machine translation to translate their own texts.

As the tools do not provide any evidence, the likelihood that an educational institution is able to prove this form of academic misconduct is extremely low. Reports provided by detection tools for AI-generated text cannot be used as the only basis for reporting students for cheating. They can give faculty a hint that some sort of misconduct may have happened, but further dialogue and conversations with students should take place.

One of the tools that the researchers came across, GLTR (http://gltr.io/) does not provide any classification, so it was decided to exclude it from testing. Nonetheless, it highlights the words (tokens) based on how commonly they appear in a given context. Interpretation of the output is up to the educator, but the research authors find the visualisation of this information very useful. The colour-coded predictability of individual words does not necessarily mean that the text was generated by AI, but may also mean that the text does not bring any innovation or added value, which might be - in some situations - a relevant indicator of its quality.

As the detection tools for AI-generated text are not reliable, a prevention-focused approach needs to be prioritised over a detection one. It is also paramount to inform the educators about this fact. The focus should instead be on the preventive pedagogical strategies on how to ethically use generative AI tools, including a discussion about the benefits and limitations of such tools.

This presupposes defining, describing, and training on the differences between the ethical and unethical use of AI tools will be important for students, faculty, and staff. The ENAI recommendations on the ethical use of Artificial Intelligence in Education may be a good starting point (Foltýnek et al., 2023) for such discussions. It is also important to encourage educators to rethink their assessment strategies and instruments to achieve a design with features that reduce or even eliminate the possibility of enabling cheating.

Our study has some limitations. It focused only on English language texts. Even though we had computer code, we did not test the performance of the systems specifically on that. There were also indications that the results from the tools can vary when the same material is tested at a different time; we did not systematically examine the replicability of the results provided by the tools. Nevertheless, we tentatively suggest that this inconsistency can have major implications in misconduct investigations and thus provides another strong reason against the use of these tools as a single source of an accusation of misconduct. Our document set is also somewhat limited: we did not test the kind of hybrid writing with iterative use of AI that may be likely to be more typical of student use of generative AI. However, the poor performance of the tools across the range of documents does not imply better performance for hybrid writing.

# 7. Conclusion and Future Work

This paper exposes serious limitations of the state-of-the-art AI-generated text detection tools and their unsuitability for use as evidence of academic misconduct. Our findings do not confirm the claims presented by the systems. They too often present false positives and false negatives. Moreover, it is too easy to game the systems by using paraphrasing tools or machine translation. Therefore, our conclusion is that the systems we tested should not be used in academic settings. Although text matching software also suffers from false positives and false negatives (Foltýnek et al., 2020), at least it is possible to provide evidence of potential misconduct. In the case of the detection tools for AI-generated text, this is not the case.

Our findings strongly suggest that the “easy solution” for detection of AI-generated text does not (and maybe even could not) exist. Therefore, rather than focusing on detection strategies, educators continue to need to focus on preventive measures and continue to rethink academic assessment strategies (see, for example, Bjelobaba 2020). Written assessment should focus on the process of development of student skills rather than the final product.

Future research in this area should test the performance of AI-generated text detection tools on texts produced with different (and multiple) levels of obfuscation e.g., the use of machine paraphrasers, translators, patch writers, etc. Another line of research might explore the detection of AI-generated text at a cohort level through its impact on student learning (e. g. through assessment scores) and education systems (e. g. the impact of generative AI on similarity scores). Research should also build on the known issues with cloud-based text-matching software to explore the legal implications and data privacy issues involved in uploading content to cloud-based (or institutional) AI detection tools.

# Declarations

## Availability of data and materials

All data and testing materials are available at https://www.academicintegrity.eu/wp/technology-academic-integrity-working-group/.

## Competing interests

Two authors of this article, SB and TF, are involved in organising the European Conference on Ethics and Integrity in Academia 2023 co-organised by the European Network for Academic Integrity. This conference receives sponsorship from Turnitin. This did not influence the research presented in the paper in any phase.

## Funding

The authors had no funding for this investigation other than from their respective institutions.

## Authors’ contributions

All authors created test data, ran the tests, collected data, discussed the statistical results, and contributed equally to the text. TF and OP prepared the statistics for discussion.

## Acknowledgements

The authors wish to thank their colleague Július Kravjar from Slovakia who contributed a full set of test documents to the investigation.

The authors also wish to thank their colleagues from Turkey, Salim Razı and Özgür Çelik, who participated in the initial stages of the discussions about this research endeavour, but due to the devastating earthquake in February 2023 were not able to contribute further.

The tool *similarity-texter* was created as part of the bachelor’s thesis of Sofia Kalaidopoulou and is based on Dick Grune’s sim_text algorithm. It was submitted to the HTW Berlin in 2016 and is available under a Creative Commons BY-NC-SA 4.0 International License at https://people.f4.htw-berlin.de/~weberwu/simtexter/app.html

ChatGPT was NOT used to tweak any portion of this publication.

# Authors' information

The authors are members of the European Network for Academic Integrity (ENAI) working group on Technology and Academic Integrity. DWW is a plagiarism researcher and a retired professor of computer science from the HTW Berlin, Germany. AAN is an associate professor at the Department of Artificial Intelligence and Systems Engineering of Riga Technical University, Latvia. SB is a researcher in research integrity at Center for Research Ethics & Bioethics, at Uppsala University, Sweden, and the Vice-president of ENAI. TF is an assistant professor at the Department of Machine Learning and Data Processing at the Faculty of Informatics, Masaryk University, Czechia, and President of ENAI. JGD is a professor of the School of Engineering from University of Monterrey, Mexico and oversees the efforts of its Center for Integrity and Ethics. OP is an Education Developer specialising in assessment integrity at Queen Mary University of London, UK. PS is a student of Computer Science at the Faculty of Informatics, Masaryk University, Czechia. LW is the University of Leeds, UK, Academic Integrity Lead.

# Appendix

## A. Case studies 05-ManEd

The following images show the generated texts on the left and the human-obfuscated ones on the right. The identical text is coloured in the same colour on both sides, with the changes popping out in white. The images were prepared using the *similarity-texter*. As can be seen, some texts were rather heavily re-written, others only had a few words exchanged.

COMPARISON OUTPUT

**FILE:** AIDT23-05-AAN-orig.txt

Search algorithms are a set of procedures used to find a particular piece of information within a data set or search space. In computer science, search algorithms are often used to solve problems such as finding the shortest path between two points, locating an item in a database, or identifying relevant web pages for a given search query.
There are many different types of search algorithms, including:
Linear search: A simple algorithm that sequentially checks each element of a data set until the desired item is found.
Binary search: A more efficient algorithm that works by dividing the search space in half at each step, thus reducing the number of elements that need to be examined.
Depth-first search: An algorithm used to traverse a graph or tree structure, exploring each path to its furthest depth before backtracking.
Breadth-first search: A similar algorithm to depth-first search, but instead explores all neighboring nodes at each level before moving on to the next level.
A* search: An algorithm that uses heuristics to guide the search towards a goal, typically used for pathfinding in computer games or robotics.
Different search algorithms have different strengths and weaknesses, and the choice of algorithm will depend on the specific problem being solved and the characteristics of the data set being searched.
Search algorithms can be applied to state space graphs to solve a variety of problems, such as finding a path from an initial state to a goal state or finding the optimal path with the lowest cost.
A state space graph represents a problem as a set of states connected by transitions or actions. The starting state is the initial state, and the goal state is the desired outcome. Search algorithms navigate this graph to find a path from the initial state to the goal state.
To apply a search algorithm to a state space graph, we typically define the following:

**FILE:** AIDT23-05-AAN-edit.txt

Search algorithms are a set of methods used to find a particular piece of data within a data set or search space. In computer science, search algorithms are often used to solve tasks such as finding the shortest path between two points, locating an item in a database, or identifying relevant web pages for a given search query.
Search algorithms are different and include:
Linear search which is a simple algorithm that sequentially checks each element of a data set until the desired item is found.
Binary search which is a more efficient algorithm that divides the search space in half at each step, thus reducing the number of elements that need to be checked.
Depth-first search which is an algorithm used to search through a graph or tree structure, exploring each path to its furthest depth before backtracking.
Breadth-first search whihc is a similar algorithm to depth-first search, but it explores all neighboring nodes at each level before moving on to the next level.
A* search which is an algorithm that uses heuristic knowledge to guide the search towards a goal; it is typically used for pathfinding in computer games or robotics.
Different search algorithms are characterized by different advantages and drawbacks, and the choice of algorithm will depend on the specific problem being solved and the characteristics of the data set being searched.
Search algorithms can be applied to state space graphs to solve a variety of problems, such as finding a path from an initial state to a goal state or finding the optimal path with the lowest cost.
A state space graph represents a problem as a set of nodes representing states connected by arcs corresponding to transitions or actions. The starting state is the initial state, and the goal state is the desired outcome. Search algorithms search through this graph to find a path from the initial state to the goal state.
To apply a search algorithm to a state space graph, typically it is necessary to define the following:

Figure A1: AIDT23-05-AAN

}

The loadDictionary() method reads in the dictionary file dictionary.txt and stores each word in a List called dictionary. The findWords() method generates all possible combinations of letters from length 1 to 7 from the given input string, and for each combination, it checks if the resulting word is in the dictionary. If it is found, the word is added to the List called words.
The program prompts the user to enter a 7-letter string, and it uses the findWords() method to find all valid words that can be created from these 7 letters. If no words are found, the program will print "No words found." If one or more words are found, the program will print "Words found:" followed by a list of the words. Note that this program assumes that the dictionary file contains one word on each line and that the words are in lowercase.

If the dictionary file does not have all words in lowercase, then it will be necessary to modify the loadDictionary() method in the Java code to convert each word in the dictionary to lowercase before adding it to the List called dictionary.
There are many online dictionaries that would be useful for a Scrabble cheater program. Here are a few options:
1. The official Scrabble dictionary: The official Scrabble dictionary contains over 100,000 words that are eligible to be played in Scrabble. The official dictionary is at the official Scrabble website.
2. Many online game sites provide word lists that can be used for cheating in their games. For example, the Words with Friends website provides a list of all the playable words in their game. You can also find word lists for other games by doing a quick search online.
3. There are many online open source dictionaries available that can be used for

The loadDictionary() method reads in the dictionary file and stores each word in a List<String> called dictionary. The findWords() method generates all possible combinations of letters of length 1 through 7 from the given input string, and for each combination, it checks if the resulting word is in the dictionary. If it is, the word is added to the List<String> called words.
The program prompts the user to enter a 7-letter string, and it uses the findWords() method to find all valid words that can be created from these letters. If no words are found, the program prints "No words found." If one or more words are found, the program prints "Words found:" followed by a list of the words. Note that this program assumes that the dictionary file contains one word per line and that the words are in lowercase.

There are many online resources where you can find dictionaries that would be useful for a Scrabble cheater program. Here are a few options:
1. The official Scrabble dictionary: The official Scrabble dictionary contains over 100,000 words that are eligible to be played in Scrabble. You can find the official dictionary at the official Scrabble website or on other online game sites.
2. Word lists from online game sites: Many online game sites provide word lists that can be used for cheating in their games. For example, the Words with Friends website provides a list of all the playable words in their game. You can also find word lists for other games, such as Scrabble or Boggle, by doing a quick search online.
3. Open source dictionaries: There are many open source dictionaries available online that can be used for free. One popular option is the Open American National Corpus (OANC), which is a large, open source corpus of American English that includes over 15 million words.
4. Paid dictionaries: There are also many paid dictionaries available online that can be used for a Scrabble cheater program. One popular option is the Official

Figure A2: AIDT23-05-DWW

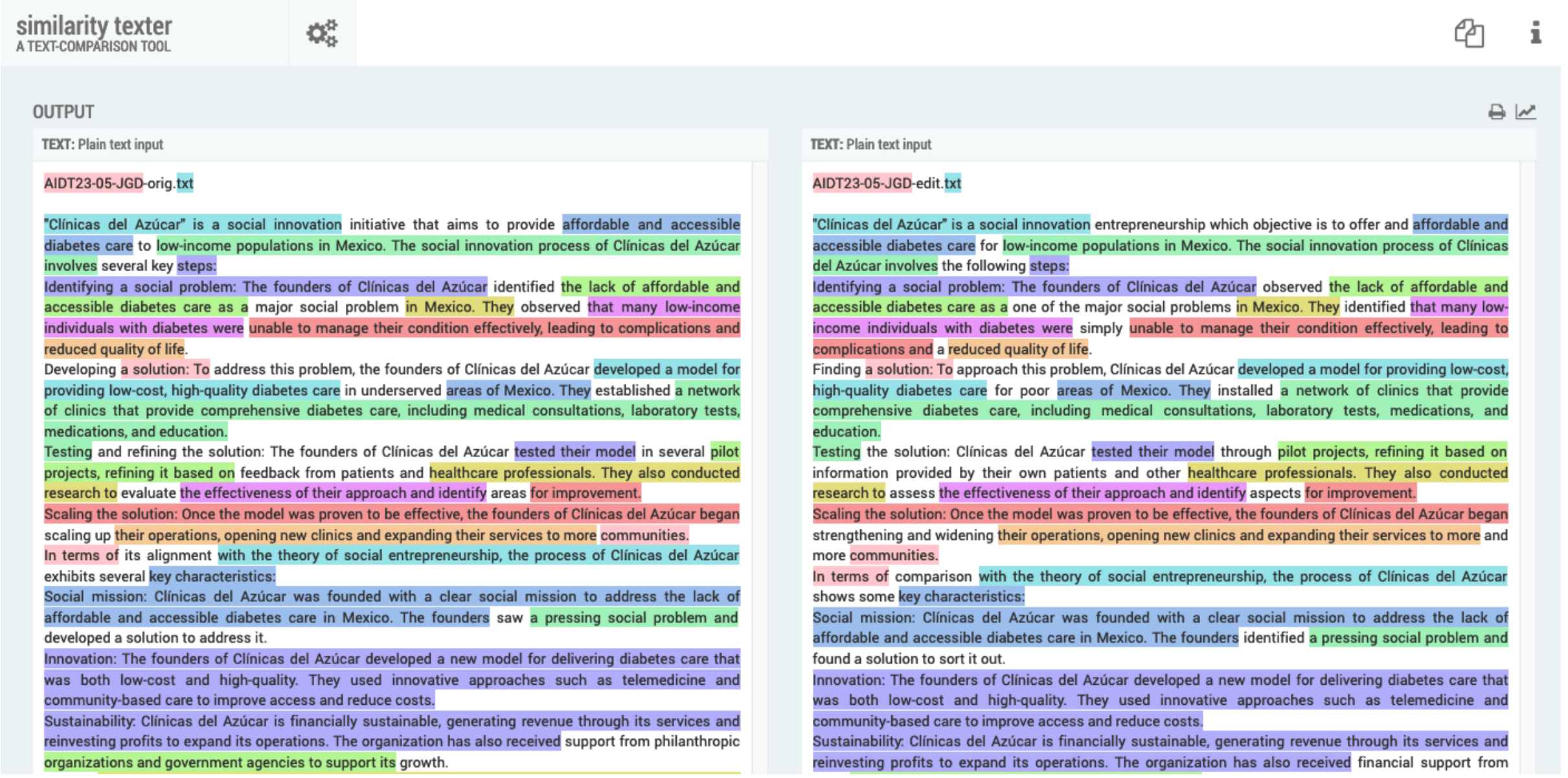
similarity texter
A TEXT-COMPARISON TOOL

OUTPUT

TEXT: Plain text input

AIDT23-05-JGD-orig.txt

"Clínicas del Azúcar" is a social innovation initiative that aims to provide affordable and accessible diabetes care to low-income populations in Mexico. The social innovation process of Clínicas del Azúcar involves several key steps:
Identifying a social problem: The founders of Clínicas del Azúcar identified the lack of affordable and accessible diabetes care as a major social problem in Mexico. They observed that many low-income individuals with diabetes were unable to manage their condition effectively, leading to complications and reduced quality of life.
Developing a solution: To address this problem, the founders of Clínicas del Azúcar developed a model for providing low-cost, high-quality diabetes care in underserved areas of Mexico. They established a network of clinics that provide comprehensive diabetes care, including medical consultations, laboratory tests, medications, and education.
Testing and refining the solution: The founders of Clínicas del Azúcar tested their model in several pilot projects, refining it based on feedback from patients and healthcare professionals. They also conducted research to evaluate the effectiveness of their approach and identify areas for improvement.
Scaling the solution: Once the model was proven to be effective, the founders of Clínicas del Azúcar began scaling up their operations, opening new clinics and expanding their services to more communities.
In terms of its alignment with the theory of social entrepreneurship, the process of Clínicas del Azúcar exhibits several key characteristics:
Social mission: Clínicas del Azúcar was founded with a clear social mission to address the lack of affordable and accessible diabetes care in Mexico. The founders saw a pressing social problem and developed a solution to address it.
Innovation: The founders of Clínicas del Azúcar developed a new model for delivering diabetes care that was both low-cost and high-quality. They used innovative approaches such as telemedicine and community-based care to improve access and reduce costs.
Sustainability: Clínicas del Azúcar is financially sustainable, generating revenue through its services and reinvesting profits to expand its operations. The organization has also received support from philanthropic organizations and government agencies to support its growth.

TEXT: Plain text input

AIDT23-05-JGD-edit.txt

"Clínicas del Azúcar" is a social innovation entrepreneurship which objective is to offer and affordable and accessible diabetes care for low-income populations in Mexico. The social innovation process of Clínicas del Azúcar involves the following steps:
Identifying a social problem: The founders of Clínicas del Azúcar observed the lack of affordable and accessible diabetes care as a one of the major social problems in Mexico. They identified that many low-income individuals with diabetes were simply unable to manage their condition effectively, leading to complications and a reduced quality of life.
Finding a solution: To approach this problem, Clínicas del Azúcar developed a model for providing low-cost, high-quality diabetes care for poor areas of Mexico. They installed a network of clinics that provide comprehensive diabetes care, including medical consultations, laboratory tests, medications, and education.
Testing the solution: Clínicas del Azúcar tested their model through pilot projects, refining it based on information provided by their own patients and other healthcare professionals. They also conducted research to assess the effectiveness of their approach and identify aspects for improvement.
Scaling the solution: Once the model was proven to be effective, the founders of Clínicas del Azúcar began strengthening and widening their operations, opening new clinics and expanding their services to more and more communities.
In terms of comparison with the theory of social entrepreneurship, the process of Clínicas del Azúcar shows some key characteristics:
Social mission: Clínicas del Azúcar was founded with a clear social mission to address the lack of affordable and accessible diabetes care in Mexico. The founders identified a pressing social problem and found a solution to sort it out.
Innovation: The founders of Clínicas del Azúcar developed a new model for delivering diabetes care that was both low-cost and high-quality. They used innovative approaches such as telemedicine and community-based care to improve access and reduce costs.
Sustainability: Clínicas del Azúcar is financially sustainable, generating revenue through its services and reinvesting profits to expand its operations. The organization has also received financial support from

Figure A3: AIDT23-05-JGD

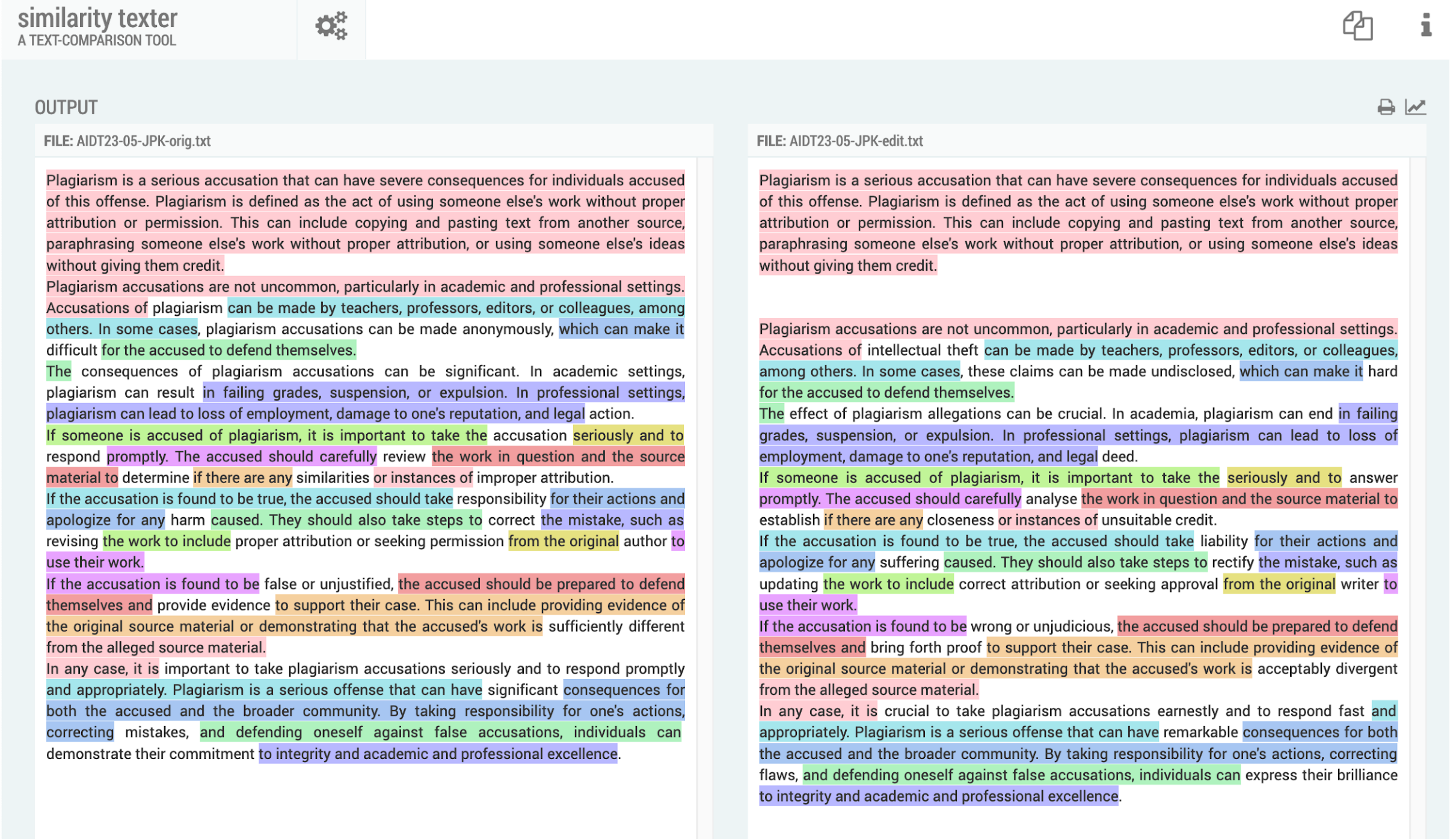

Figure A4: AIDT23-05-JPK

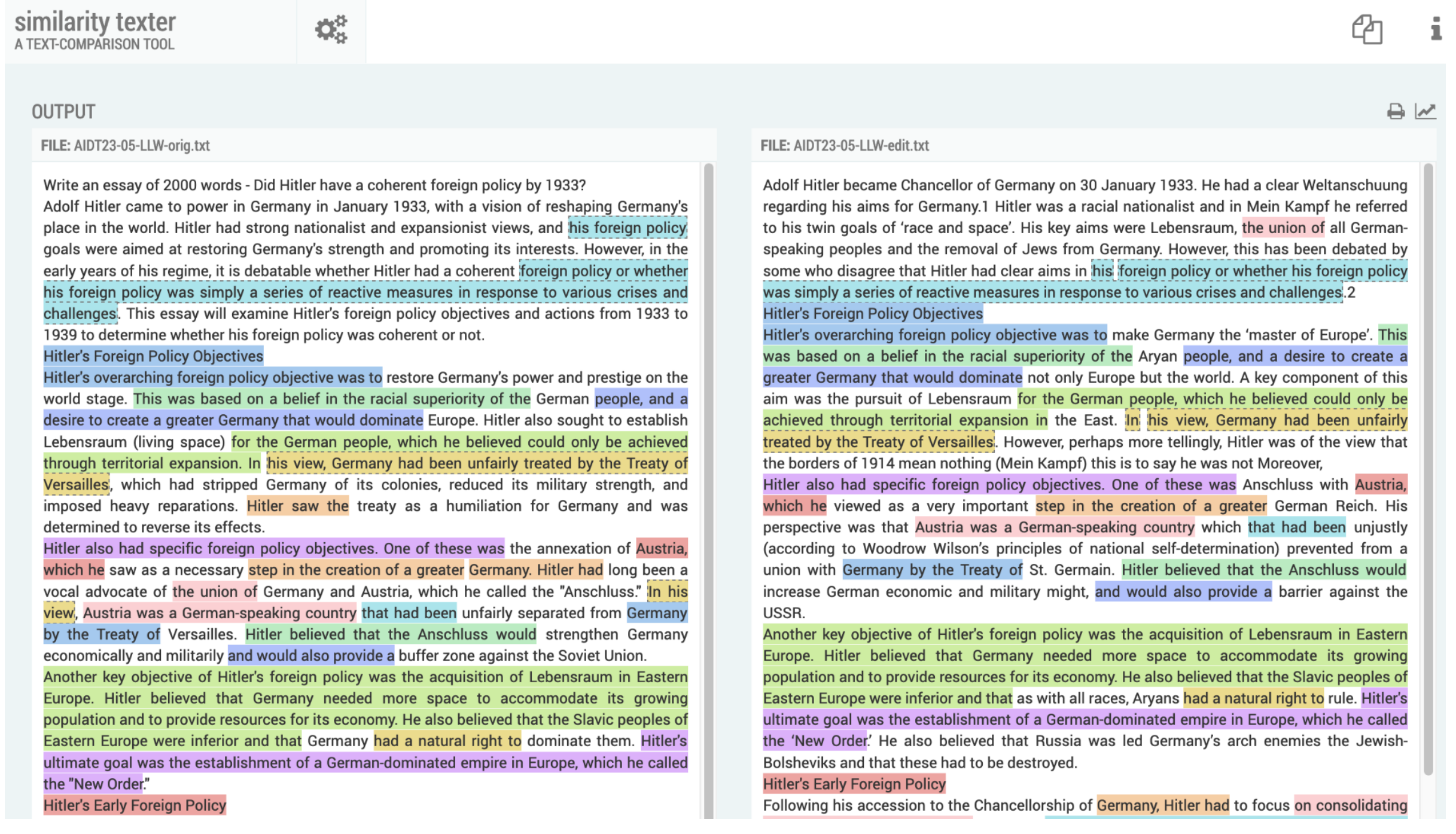

Figure A5: AIDT23-05-LLW

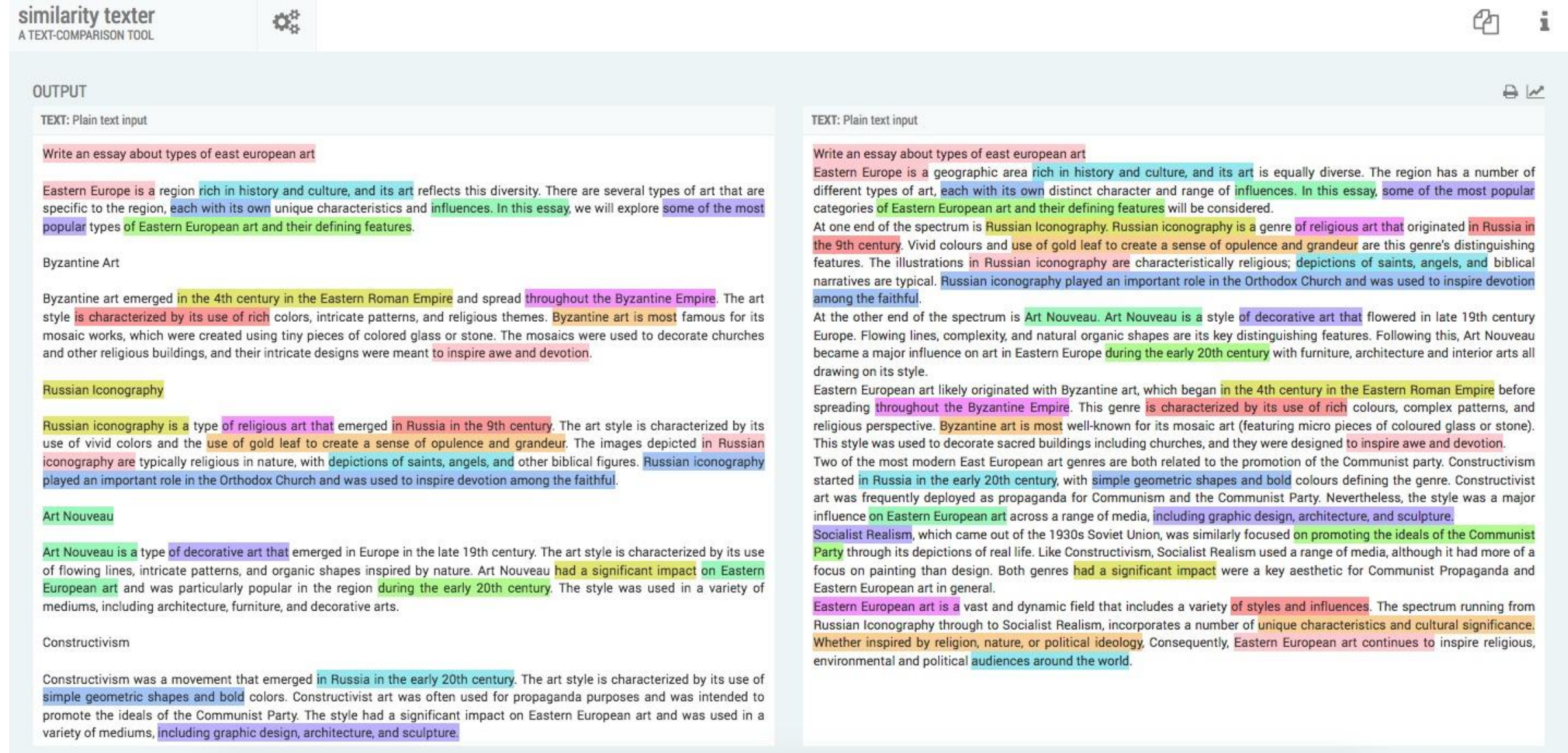

Figure A6: AIDT23-05-OLU

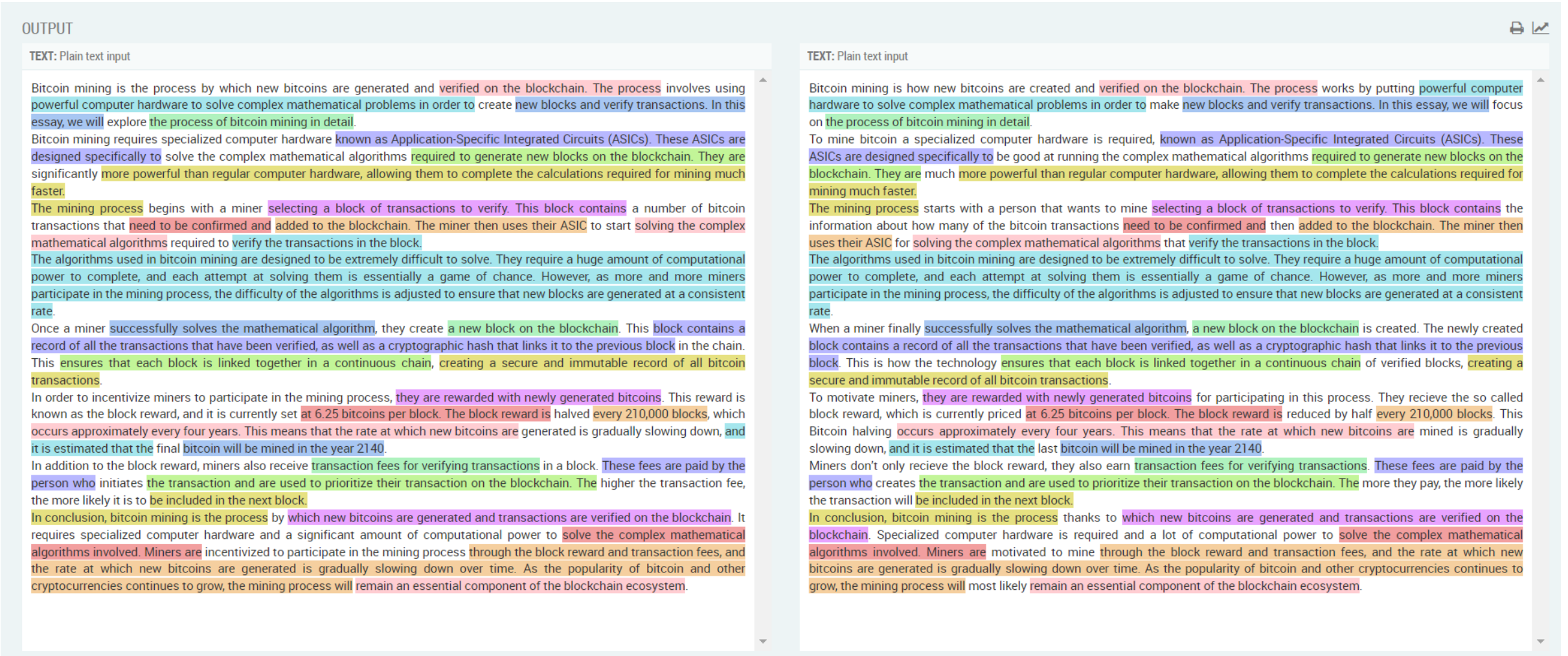

Figure A7: AIDT23-05-PTR

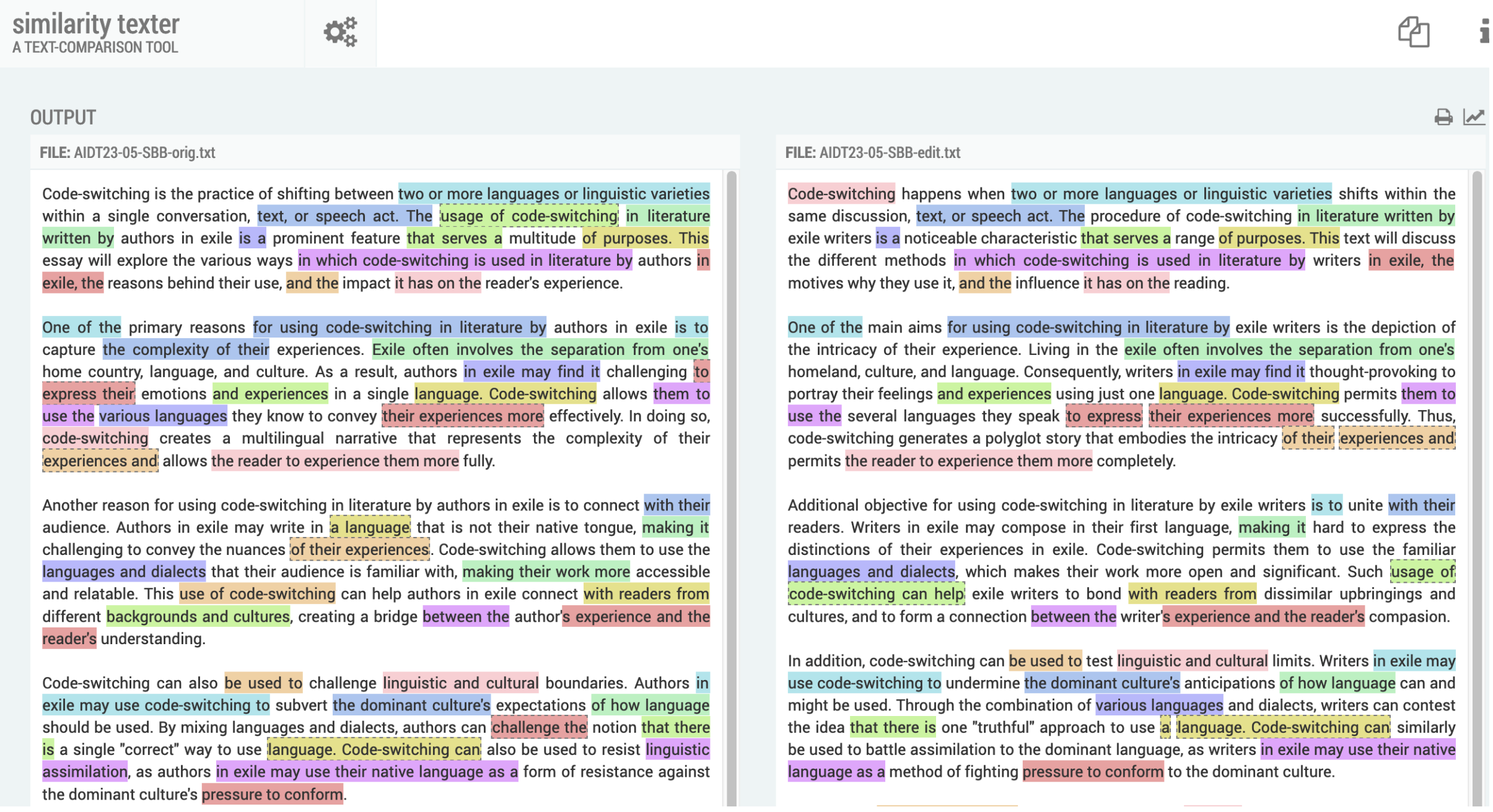

similarity texter
A TEXT-COMPARISON TOOL

OUTPUT

**FILE:** AIDT23-05-SBB-orig.txt

Code-switching is the practice of shifting between two or more languages or linguistic varieties within a single conversation, text, or speech act. The usage of code-switching in literature written by authors in exile is a prominent feature that serves a multitude of purposes. This essay will explore the various ways in which code-switching is used in literature by authors in exile, the reasons behind their use, and the impact it has on the reader's experience.

One of the primary reasons for using code-switching in literature by authors in exile is to capture the complexity of their experiences. Exile often involves the separation from one's home country, language, and culture. As a result, authors in exile may find it challenging to express their emotions and experiences in a single language. Code-switching allows them to use the various languages they know to convey their experiences more effectively. In doing so, code-switching creates a multilingual narrative that represents the complexity of their experiences and allows the reader to experience them more fully.

Another reason for using code-switching in literature by authors in exile is to connect with their audience. Authors in exile may write in a language that is not their native tongue, making it challenging to convey the nuances of their experiences. Code-switching allows them to use the languages and dialects that their audience is familiar with, making their work more accessible and relatable. This use of code-switching can help authors in exile connect with readers from different backgrounds and cultures, creating a bridge between the author's experience and the reader's understanding.

Code-switching can also be used to challenge linguistic and cultural boundaries. Authors in exile may use code-switching to subvert the dominant culture's expectations of how language should be used. By mixing languages and dialects, authors can challenge the notion that there is a single "correct" way to use language. Code-switching can also be used to resist linguistic assimilation, as authors in exile may use their native language as a form of resistance against the dominant culture's pressure to conform.

**FILE:** AIDT23-05-SBB-edit.txt

Code-switching happens when two or more languages or linguistic varieties shifts within the same discussion, text, or speech act. The procedure of code-switching in literature written by exile writers is a noticeable characteristic that serves a range of purposes. This text will discuss the different methods in which code-switching is used in literature by writers in exile, the motives why they use it, and the influence it has on the reading.

One of the main aims for using code-switching in literature by exile writers is the depiction of the intricacy of their experience. Living in the exile often involves the separation from one's homeland, culture, and language. Consequently, writers in exile may find it thought-provoking to portray their feelings and experiences using just one language. Code-switching permits them to use the several languages they speak to express their experiences more successfully. Thus, code-switching generates a polyglot story that embodies the intricacy of their experiences and permits the reader to experience them more completely.

Additional objective for using code-switching in literature by exile writers is to unite with their readers. Writers in exile may compose in their first language, making it hard to express the distinctions of their experiences in exile. Code-switching permits them to use the familiar languages and dialects, which makes their work more open and significant. Such usage of code-switching can help exile writers to bond with readers from dissimilar upbringings and cultures, and to form a connection between the writer's experience and the reader's compasion.

In addition, code-switching can be used to test linguistic and cultural limits. Writers in exile may use code-switching to undermine the dominant culture's anticipations of how language can and might be used. Through the combination of various languages and dialects, writers can contest the idea that there is one "truthful" approach to use a language. Code-switching can similarly be used to battle assimilation to the dominant language, as writers in exile may use their native language as a method of fighting pressure to conform to the dominant culture.

Figure A8: AIDT23-05-SBB

## COMPARISON OUTPUT

**FILE:** AIDT23-05-TFO-orig.txt

Website blocking is a contentious issue that raises many questions about freedom of speech and protection of democratic values. Recently, in response to the Russian invasion of Ukraine, some websites containing Russian propaganda were blocked. While some argue that website blocking is necessary to prevent the spread of false information and protect national security, others contend that it violates the principles of freedom of speech and can lead to censorship. In this essay, we will explore both the pros and cons of website blocking.

On the one hand, website blocking can be viewed as a necessary measure to protect national security and prevent the spread of false information. In the case of the Russian propaganda websites, blocking them can help to prevent the dissemination of misinformation that can harm the public and national interest. It can also serve as a warning to other countries that they will not be able to use the internet as a tool to manipulate other nations. In this way, website blocking can be seen as a proactive step in safeguarding democratic values.

On the other hand, website blocking can be viewed as a violation of the principles of freedom of speech and a form of censorship. Free speech is a fundamental principle of democracy, and website blocking can be seen as a threat to this principle. When a government blocks a website, it is essentially restricting access to information, and this can lead to a lack of transparency and accountability. In addition, website blocking can be seen as a slippery slope towards increased censorship and government control over the internet.

**FILE:** AIDT23-05-TFO-edit.txt

Website blocking is a sensitive issue that raises many questions about freedom of speech and protection of democratic values. Recently, in response to the Russian invasion of Ukraine, some websites containing Russian propaganda were blocked. While some argue that website blocking is necessary to prevent the spread of disinformation and protect national security, others contend that it violates the principles of freedom of speech and can lead to censorship. In this essay, we will explore both the pros and cons of website blocking.

On the one hand, website blocking can be viewed as a necessary measure to protect national security and prevent the spread of disinformation. In the case of the Russian propaganda websites, blocking them can help to prevent the dissemination of disinformation that can harm the public and national interest. It can also serve as a warning to other countries that they will not be able to use the internet as a tool to manipulate other nations. In this way, website blocking can be seen as a proactive step in safeguarding democratic values.

On the other hand, website blocking can be viewed as a violation of the principles of freedom of speech and a form of censorship. Free speech is a fundamental principle of democracy, and website blocking is certainly a threat to this principle. When a government blocks a website, it is essentially restricting access to some opinions, and this can lead to a lack of transparency and accountability. In addition, some see website blocking as a slippery slope towards increased censorship and government control over the internet.

Figure A9: AIDT23-05-TFO

# B. Case studies 06-Para

These test cases were first generated with ChatGPT, then automatically re-written using Quillbot with the default settings. The generated original is on the left, the re-written version on the right.

## COMPARISON OUTPUT

**FILE:** AIDT23-06-AAN-orig.txt

A pipeline in machine learning is a sequence of data processing components, or steps, that are chained together in a specific order to execute a complete machine learning workflow.

A typical pipeline in machine learning involves the following stages:

Data Collection: The first stage of any machine learning pipeline is to gather the data required for training the machine learning model. This can involve collecting data from a variety of sources, including databases, APIs, or other online resources.

Data Cleaning and Preprocessing: Once the data is collected, it needs to be cleaned and preprocessed. This stage includes removing any irrelevant or redundant data, filling in missing values, scaling, normalization, and other techniques to transform the data into a suitable format for machine learning.

Feature Extraction and Selection: The next stage is to identify the features or attributes that are most important for the model to learn. This involves selecting relevant variables, transforming data, and reducing the dimensionality of the data if required.

Model Training: The model training stage involves feeding the preprocessed data into a chosen machine learning algorithm to build a predictive model. The algorithm learns patterns and relationships between the data and labels using various optimization techniques.

Model Evaluation: Once the model is trained, it needs to be evaluated on a separate dataset to determine its accuracy and effectiveness. This stage involves metrics like accuracy, precision, recall, F1 score, AUC-ROC, etc.

Model Deployment: Finally, the trained model is deployed into a production environment to make predictions on new data.

Overall, a pipeline in machine learning helps to automate and streamline the entire machine learning workflow, from data collection to model deployment, and makes it more efficient and scalable.

The most challenging steps in a machine learning pipeline can vary depending on the specific problem and dataset. However, some of the most challenging steps are:

Data Collection: Gathering high-quality data can be a significant challenge in many cases. Issues such as data availability, quality, and diversity can significantly impact the performance of the model.

**FILE:** AIDT23-06-AAN-edit.txt

In machine learning, a pipeline is a series of data processing processes or components that are linked together in a precise order to carry out an entire machine learning workflow.

The phases below make up a typical machine learning pipeline:

Data Gathering The gathering of the data necessary for the machine learning model's training is the initial step in any machine learning pipeline. This may entail gathering information from a number of databases, APIs, or other web resources.

Data Preprocessing and Cleaning Data needs to be cleansed and preprocessed after it is obtained. At this stage, the data is transformed into a format that is suitable for machine learning by deleting any redundant or irrelevant information, adding any missing values, scaling the data, and other methods.

Extraction and Selection of Features: The next step is to determine which features or attributes are crucial for the model to learn. This entails choosing pertinent variables, transforming the data, and, if necessary, lowering its dimensionality.

Model Training: In order to create a predictive model, a machine learning algorithm is fed the preprocessed data during the model training step. By the use of multiple optimization strategies, the algorithm discovers patterns and connections between the data and labels.

Model evaluation: After the model has been trained, it must be assessed on a different dataset to ascertain its precision and efficacy. Metrics including recall, accuracy, precision, F1 score, AUC-ROC, and others are used in this stage.

Model Deployment: The trained model is then put into use in a real-world setting to generate predictions about fresh data.

Ultimately, a machine learning pipeline makes it possible to automate, streamline, and scale the whole machine learning workflow, from data collection to model deployment.

Depending on the particular problem and dataset, a machine learning pipeline's most difficult steps can change. The following are some of the most difficult steps:

Data Gathering In many situations, gathering high-quality data can be very difficult. The model's performance can be severely impacted by problems with data diversity, quality, and availability.

Figure B1: AIDT23-06-AAN

## COMPARISON OUTPUT

**FILE: AIDT23-06-DWW-edit.txt**

By incorporating structured and machine-readable data into web pages, the Semantic Web is an expansion of the World Wide Web that intends to give users a richer and more meaningful experience. The Semantic Web is an endeavor to improve the machine understanding of web content so that computers can better grasp the relationships between various kinds of data and give users more relevant and customized results.

The Semantic Web is founded on the notion that web material may be represented and organized using ontologies, which are formal definitions of concepts and their relationships. By doing this, it is able to link several bits of online data together and reason more intelligently and automatically about them.

In the Semantic Web, RDF (Resource Description Framework), OWL (Web Ontology Language), and SPARQL are some of the important technologies (a query language for RDF data). These tools give programmers the ability to build ontologies, utilize them to describe data, and query and manage that data in a more organized and consistent manner.
In general, by enabling machines to understand and reason about web content better, the Semantic Web holds the promise of making the web more intelligent and valuable for consumers.

The Semantic Web is a genuine and active area of research and development in computer science and related subjects, not merely a theory or an abstract idea. Since Tim Berners-Lee, the creator of the World Wide Web, originally put forth the concept of the Semantic Web in a paper he released in 2001, there have been several attempts to implement and deploy Semantic Web technologies.

Semantic Web technologies are already being used by many large organizations, including Google, Yahoo, and the BBC, to enhance their search results and offer more individualized information to their consumers. The Linked Data movement and the Horizon 2020 program of the European Union are only two examples of the numerous public and corporate initiatives that aim to build large-scale, interconnected networks of data utilizing Semantic Web technology.

**FILE: AIDT23-06-DWW-orig.txt**

The Semantic Web is an extension of the World Wide Web that aims to provide a richer and more meaningful experience for users by adding structured and machine-readable data to web pages. In essence, the Semantic Web is an effort to make web content more understandable by machines, so that they can better understand the relationships between different pieces of data and provide more relevant and personalized results to users.
The Semantic Web is based on the idea of using ontologies, which are formal descriptions of concepts and their relationships, to represent and structure web content. By doing so, it becomes possible to create links between different pieces of data on the web and to reason about them in a more intelligent and automated way.
Some of the key technologies used in the Semantic Web include RDF (Resource Description Framework), OWL (Web Ontology Language), and SPARQL (a query language for RDF data). These technologies enable developers to create and use ontologies to represent data and to query and manipulate that data in a more structured and consistent way.
Overall, the Semantic Web holds the promise of making the web more intelligent and more useful for users, by enabling machines to better understand and reason about web content.

The Semantic Web is not just a theory or an abstract concept, but it is a real and active area of research and development in computer science and related fields. The idea of the Semantic Web was first proposed by Tim Berners-Lee, the inventor of the World Wide Web, in a paper he published in 2001, and since then, there have been numerous efforts to implement and deploy Semantic Web technologies.
Many large organizations, such as Google, Yahoo, and the BBC, have already started using Semantic Web technologies to improve their search results and to provide more personalized content to their users. In addition, there are many public and private initiatives that aim to create large-scale, interconnected networks of data using Semantic Web technologies, such as the Linked Data movement and the European Union's Horizon 2020 program.
While the Semantic Web is still a relatively new and evolving field, there is a growing body of research and real-world examples that demonstrate the potential of these technologies to improve the web experience for both users

Figure B2: AIDT23-06-DWW

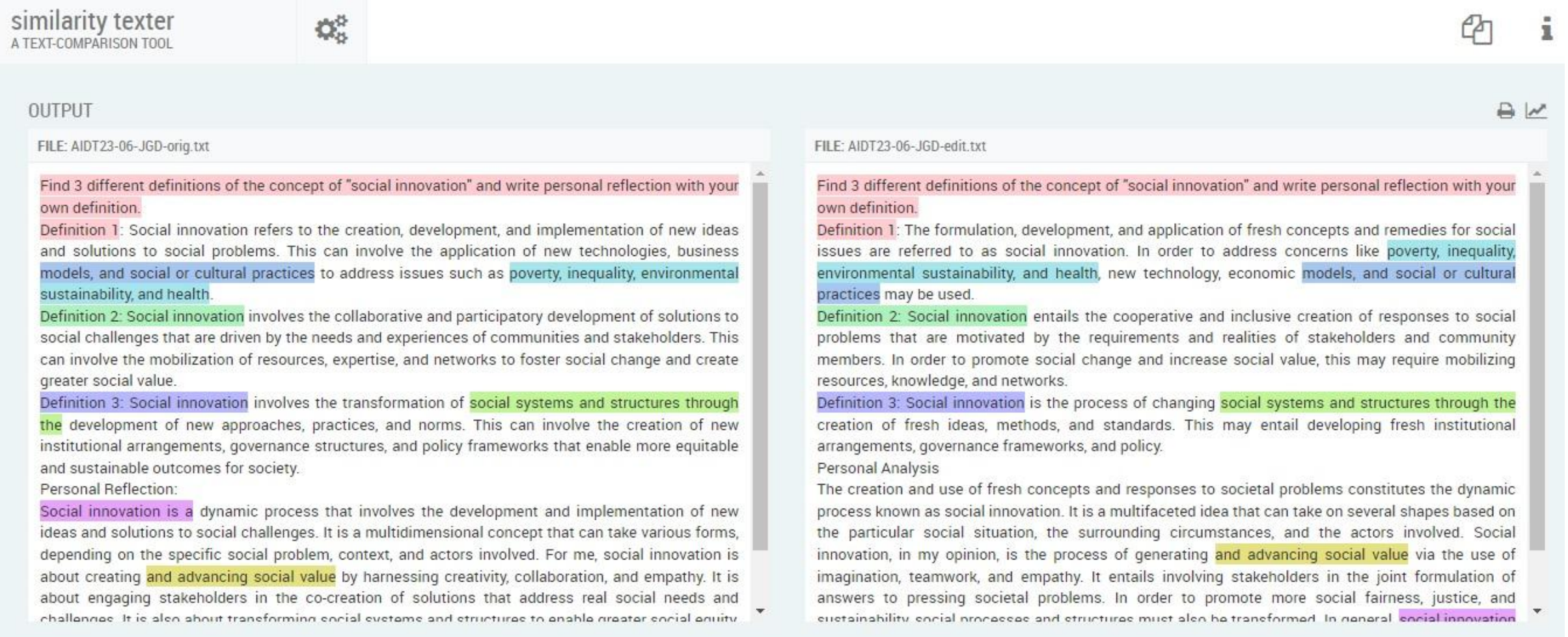
similarity texter

A TEXT-COMPARISON TOOL

OUTPUT

FILE: AIDT23-06-JGD-orig.txt

Find 3 different definitions of the concept of "social innovation" and write personal reflection with your own definition.
Definition 1: Social innovation refers to the creation, development, and implementation of new ideas and solutions to social problems. This can involve the application of new technologies, business models, and social or cultural practices to address issues such as poverty, inequality, environmental sustainability, and health.
Definition 2: Social innovation involves the collaborative and participatory development of solutions to social challenges that are driven by the needs and experiences of communities and stakeholders. This can involve the mobilization of resources, expertise, and networks to foster social change and create greater social value.
Definition 3: Social innovation involves the transformation of social systems and structures through the development of new approaches, practices, and norms. This can involve the creation of new institutional arrangements, governance structures, and policy frameworks that enable more equitable and sustainable outcomes for society.
Personal Reflection:
Social innovation is a dynamic process that involves the development and implementation of new ideas and solutions to social challenges. It is a multidimensional concept that can take various forms, depending on the specific social problem, context, and actors involved. For me, social innovation is about creating and advancing social value by harnessing creativity, collaboration, and empathy. It is about engaging stakeholders in the co-creation of solutions that address real social needs and

FILE: AIDT23-06-JGD-edit.txt

Find 3 different definitions of the concept of "social innovation" and write personal reflection with your own definition.
Definition 1: The formulation, development, and application of fresh concepts and remedies for social issues are referred to as social innovation. In order to address concerns like poverty, inequality, environmental sustainability, and health, new technology, economic models, and social or cultural practices may be used.
Definition 2: Social innovation entails the cooperative and inclusive creation of responses to social problems that are motivated by the requirements and realities of stakeholders and community members. In order to promote social change and increase social value, this may require mobilizing resources, knowledge, and networks.
Definition 3: Social innovation is the process of changing social systems and structures through the creation of fresh ideas, methods, and standards. This may entail developing fresh institutional arrangements, governance frameworks, and policy.
Personal Analysis
The creation and use of fresh concepts and responses to societal problems constitutes the dynamic process known as social innovation. It is a multifaceted idea that can take on several shapes based on the particular social situation, the surrounding circumstances, and the actors involved. Social innovation, in my opinion, is the process of generating and advancing social value via the use of imagination, teamwork, and empathy. It entails involving stakeholders in the joint formulation of answers to pressing societal problems. In order to promote more social fairness, justice, and

Figure B3: AIDT23-06-JGD

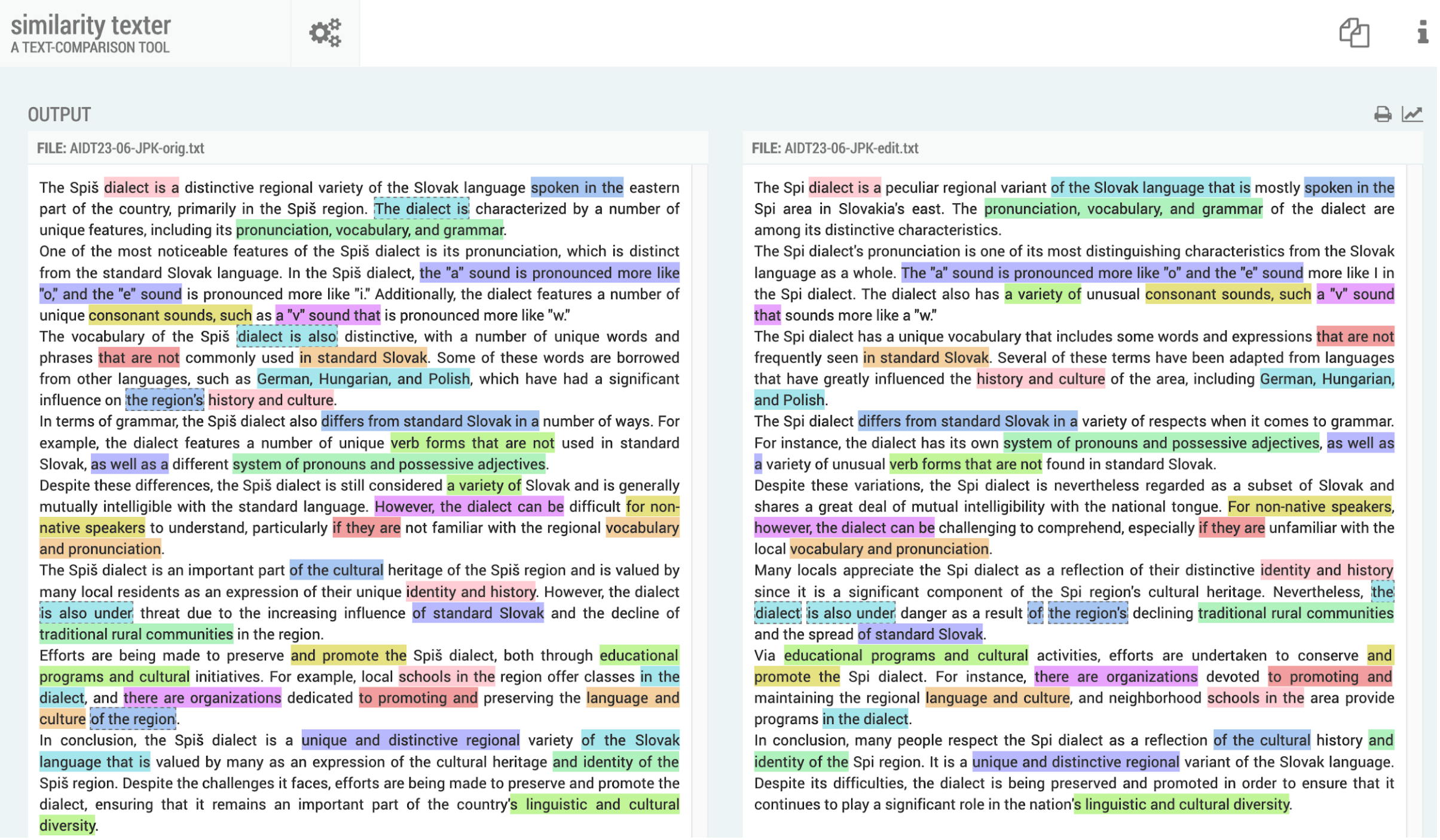
similarity texter
A TEXT-COMPARISON TOOL

OUTPUT

FILE: AIDT23-06-JPK-orig.txt

The Spiš dialect is a distinctive regional variety of the Slovak language spoken in the eastern part of the country, primarily in the Spiš region. The dialect is characterized by a number of unique features, including its pronunciation, vocabulary, and grammar.
One of the most noticeable features of the Spiš dialect is its pronunciation, which is distinct from the standard Slovak language. In the Spiš dialect, the "a" sound is pronounced more like "o," and the "e" sound is pronounced more like "i." Additionally, the dialect features a number of unique consonant sounds, such as a "v" sound that is pronounced more like "w."
The vocabulary of the Spiš dialect is also distinctive, with a number of unique words and phrases that are not commonly used in standard Slovak. Some of these words are borrowed from other languages, such as German, Hungarian, and Polish, which have had a significant influence on the region's history and culture.
In terms of grammar, the Spiš dialect also differs from standard Slovak in a number of ways. For example, the dialect features a number of unique verb forms that are not used in standard Slovak, as well as a different system of pronouns and possessive adjectives.
Despite these differences, the Spiš dialect is still considered a variety of Slovak and is generally mutually intelligible with the standard language. However, the dialect can be difficult for non-native speakers to understand, particularly if they are not familiar with the regional vocabulary and pronunciation.
The Spiš dialect is an important part of the cultural heritage of the Spiš region and is valued by many local residents as an expression of their unique identity and history. However, the dialect is also under threat due to the increasing influence of standard Slovak and the decline of traditional rural communities in the region.
Efforts are being made to preserve and promote the Spiš dialect, both through educational programs and cultural initiatives. For example, local schools in the region offer classes in the dialect, and there are organizations dedicated to promoting and preserving the language and culture of the region.
In conclusion, the Spiš dialect is a unique and distinctive regional variety of the Slovak language that is valued by many as an expression of the cultural heritage and identity of the Spiš region. Despite the challenges it faces, efforts are being made to preserve and promote the dialect, ensuring that it remains an important part of the country's linguistic and cultural diversity.

FILE: AIDT23-06-JPK-edit.txt

The Spi dialect is a peculiar regional variant of the Slovak language that is mostly spoken in the Spi area in Slovakia's east. The pronunciation, vocabulary, and grammar of the dialect are among its distinctive characteristics.
The Spi dialect's pronunciation is one of its most distinguishing characteristics from the Slovak language as a whole. The "a" sound is pronounced more like "o" and the "e" sound more like I in the Spi dialect. The dialect also has a variety of unusual consonant sounds, such a "v" sound that sounds more like a "w."
The Spi dialect has a unique vocabulary that includes some words and expressions that are not frequently seen in standard Slovak. Several of these terms have been adapted from languages that have greatly influenced the history and culture of the area, including German, Hungarian, and Polish.
The Spi dialect differs from standard Slovak in a variety of respects when it comes to grammar. For instance, the dialect has its own system of pronouns and possessive adjectives, as well as a variety of unusual verb forms that are not found in standard Slovak.
Despite these variations, the Spi dialect is nevertheless regarded as a subset of Slovak and shares a great deal of mutual intelligibility with the national tongue. For non-native speakers, however, the dialect can be challenging to comprehend, especially if they are unfamiliar with the local vocabulary and pronunciation.
Many locals appreciate the Spi dialect as a reflection of their distinctive identity and history since it is a significant component of the Spi region's cultural heritage. Nevertheless, the dialect is also under danger as a result of the region's declining traditional rural communities and the spread of standard Slovak.
Via educational programs and cultural activities, efforts are undertaken to conserve and promote the Spi dialect. For instance, there are organizations devoted to promoting and maintaining the regional language and culture, and neighborhood schools in the area provide programs in the dialect.
In conclusion, many people respect the Spi dialect as a reflection of the cultural history and identity of the Spi region. It is a unique and distinctive regional variant of the Slovak language. Despite its difficulties, the dialect is being preserved and promoted in order to ensure that it continues to play a significant role in the nation's linguistic and cultural diversity.

Figure B4: AIDT23-06-JPK

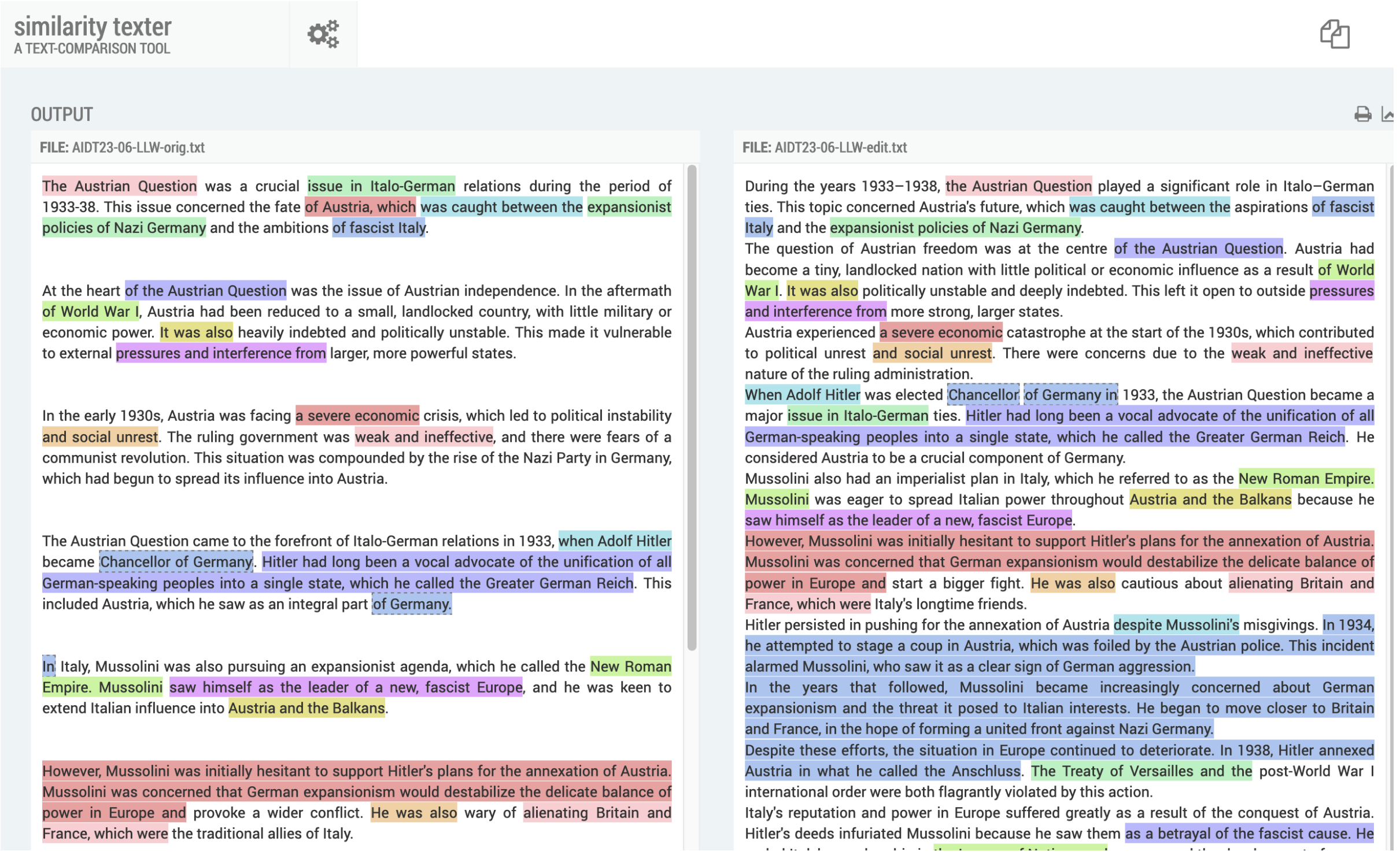
similarity texter
A TEXT-COMPARISON TOOL

OUTPUT

FILE: AIDT23-06-LLW-orig.txt

The Austrian Question was a crucial issue in Italo-German relations during the period of 1933-38. This issue concerned the fate of Austria, which was caught between the expansionist policies of Nazi Germany and the ambitions of fascist Italy.

At the heart of the Austrian Question was the issue of Austrian independence. In the aftermath of World War I, Austria had been reduced to a small, landlocked country, with little military or economic power. It was also heavily indebted and politically unstable. This made it vulnerable to external pressures and interference from larger, more powerful states.

In the early 1930s, Austria was facing a severe economic crisis, which led to political instability and social unrest. The ruling government was weak and ineffective, and there were fears of a communist revolution. This situation was compounded by the rise of the Nazi Party in Germany, which had begun to spread its influence into Austria.

The Austrian Question came to the forefront of Italo-German relations in 1933, when Adolf Hitler became Chancellor of Germany. Hitler had long been a vocal advocate of the unification of all German-speaking peoples into a single state, which he called the Greater German Reich. This included Austria, which he saw as an integral part of Germany.

In Italy, Mussolini was also pursuing an expansionist agenda, which he called the New Roman Empire. Mussolini saw himself as the leader of a new, fascist Europe, and he was keen to extend Italian influence into Austria and the Balkans.

However, Mussolini was initially hesitant to support Hitler's plans for the annexation of Austria. Mussolini was concerned that German expansionism would destabilize the delicate balance of power in Europe and provoke a wider conflict. He was also wary of alienating Britain and France, which were the traditional allies of Italy.

FILE: AIDT23-06-LLW-edit.txt

During the years 1933–1938, the Austrian Question played a significant role in Italo–German ties. This topic concerned Austria's future, which was caught between the aspirations of fascist Italy and the expansionist policies of Nazi Germany.
The question of Austrian freedom was at the centre of the Austrian Question. Austria had become a tiny, landlocked nation with little political or economic influence as a result of World War I. It was also politically unstable and deeply indebted. This left it open to outside pressures and interference from more strong, larger states.
Austria experienced a severe economic catastrophe at the start of the 1930s, which contributed to political unrest and social unrest. There were concerns due to the weak and ineffective nature of the ruling administration.
When Adolf Hitler was elected Chancellor of Germany in 1933, the Austrian Question became a major issue in Italo-German ties. Hitler had long been a vocal advocate of the unification of all German-speaking peoples into a single state, which he called the Greater German Reich. He considered Austria to be a crucial component of Germany.
Mussolini also had an imperialist plan in Italy, which he referred to as the New Roman Empire. Mussolini was eager to spread Italian power throughout Austria and the Balkans because he saw himself as the leader of a new, fascist Europe.
However, Mussolini was initially hesitant to support Hitler's plans for the annexation of Austria. Mussolini was concerned that German expansionism would destabilize the delicate balance of power in Europe and start a bigger fight. He was also cautious about alienating Britain and France, which were Italy's longtime friends.
Hitler persisted in pushing for the annexation of Austria despite Mussolini's misgivings. In 1934, he attempted to stage a coup in Austria, which was foiled by the Austrian police. This incident alarmed Mussolini, who saw it as a clear sign of German aggression.
In the years that followed, Mussolini became increasingly concerned about German expansionism and the threat it posed to Italian interests. He began to move closer to Britain and France, in the hope of forming a united front against Nazi Germany.
Despite these efforts, the situation in Europe continued to deteriorate. In 1938, Hitler annexed Austria in what he called the Anschluss. The Treaty of Versailles and the post-World War I international order were both flagrantly violated by this action.
Italy's reputation and power in Europe suffered greatly as a result of the conquest of Austria. Hitler's deeds infuriated Mussolini because he saw them as a betrayal of the fascist cause. He

Figure B5: AIDT23-06-LLW

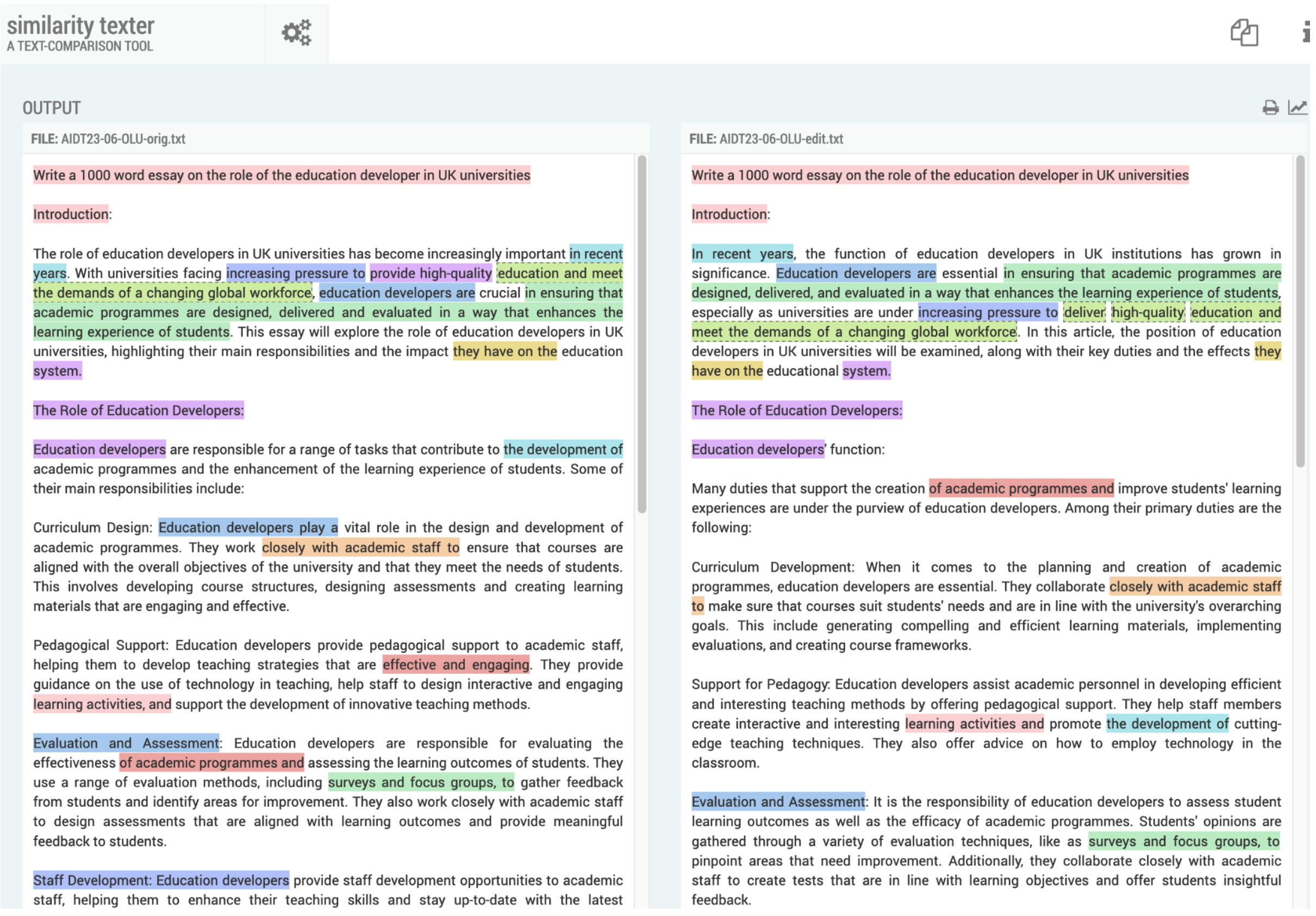

Figure B6: AIDT23-06-OLU

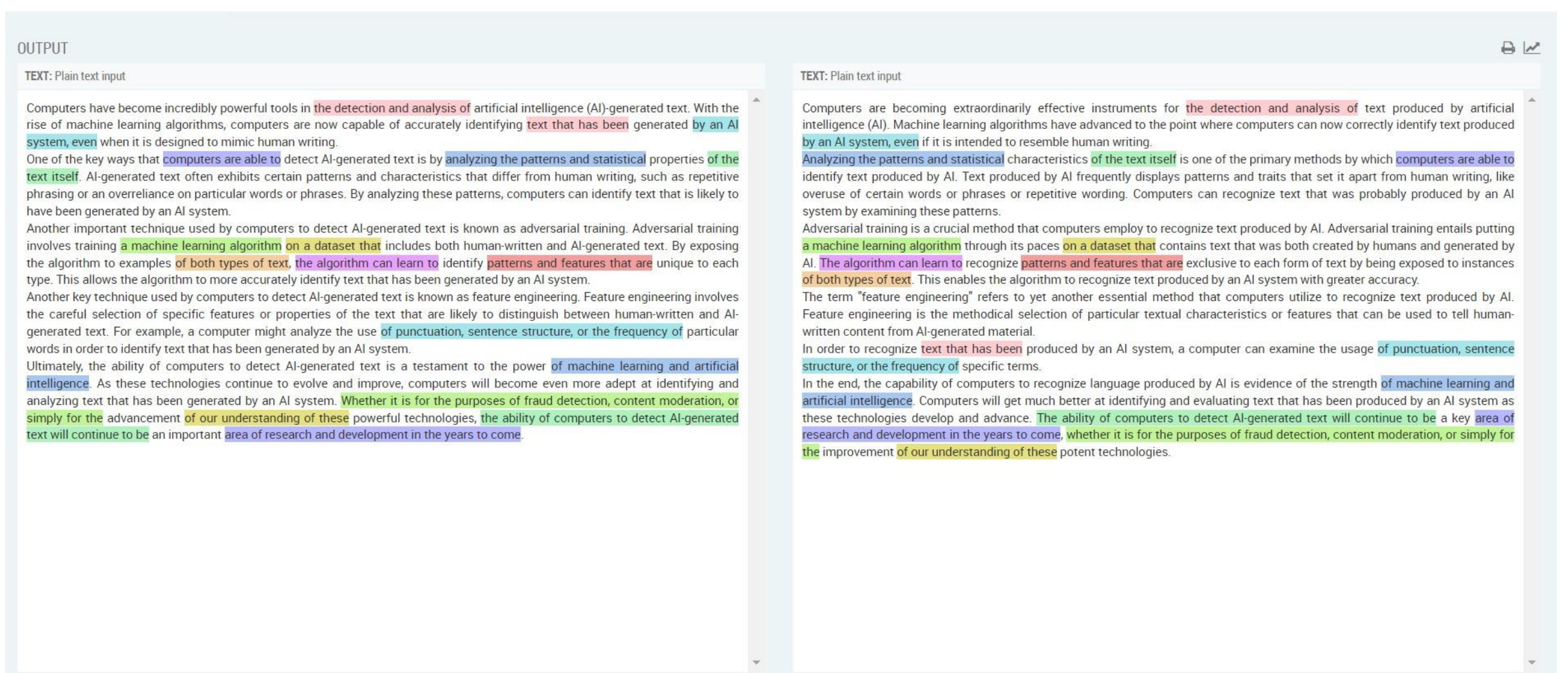

Figure B7: AIDT23-06-PTR

similarity texter
A TEXT-COMPARISON TOOL

OUTPUT

FILE: AIDT23-06-SBB-orig.txt

The disintegration of Yugoslavia in the early 1990s resulted in the emergence of several independent nation-states in the Balkan region. One of the most significant outcomes of this process was the rise of postsocialist nationalism, which played a pivotal role in shaping the political, social, and cultural dynamics of the region. This essay will explore the concept of postsocialist nationalism in the former Yugoslavia, examining its key features, factors that contributed to its emergence, and its impact on the region.

Postsocialist nationalism can be defined as a political ideology that emerged in the aftermath of the collapse of the socialist regimes in Eastern Europe, including Yugoslavia. It is characterized by a sense of national pride, a desire for self-determination, and a rejection of external interference or influence. In the case of the former Yugoslavia, postsocialist nationalism manifested itself in the form of ethno-nationalism, which prioritized the interests of individual ethnic groups over those of the state as a whole.

The roots of postsocialist nationalism in the former Yugoslavia can be traced back to the country's complex history, which was marked by centuries of ethnic and religious conflict. Under the socialist regime led by Josip Broz Tito, however, the country enjoyed a period of relative stability and prosperity, based on a model of socialist federalism that aimed to balance the interests of different ethnic groups. This system, however, began to unravel in the 1980s, as economic stagnation and political corruption undermined the legitimacy of the regime, and the death of Tito in 1980 created a power vacuum that was filled by nationalist forces.

The emergence of postsocialist nationalism in the former Yugoslavia was also shaped by broader geopolitical trends, including the collapse of the Soviet Union, the end of the Cold War, and the rise of neoliberalism. These developments created a new international environment that encouraged the reassertion of national identities and the pursuit of economic and political sovereignty. In the context of the Balkans, this trend was particularly pronounced, as different ethnic groups began to assert their claims to territory, resources, and political power.

The impact of postsocialist nationalism on the former Yugoslavia was profound, and its legacy continues to shape the region today. The disintegration of Yugoslavia was marked by a series of violent conflicts, including the war in Bosnia-Herzegovina and the Kosovo War, which resulted in the deaths of hundreds of thousands of people and the displacement of millions. These conflicts were driven by competing nationalist narratives, which portrayed different ethnic groups as victims and justified acts of violence as necessary for self-defense.

FILE: AIDT23-06-SBB-edit.txt

Several separate nation-states appeared in the Balkan region as a result of Yugoslavia's breakup in the early 1990s. The growth of postsocialist nationalism, which had a huge impact on the political, social, and cultural dynamics of the area, was one of this process' most important effects. The idea of postsocialist nationalism in the former Yugoslavia will be explored in this essay, along with its main characteristics, causes, and regional effects.

A political philosophy known as postsocialist nationalism, which included Yugoslavia, formed in the wake of the fall of socialist regimes in Eastern Europe, can be described as such. National pride, a desire for autonomy, and a rejection of outside meddling or influence are its defining characteristics. In the case of the former Yugoslavia, postsocialist nationalism took the shape of ethno-nationalism, which gave particular ethnic groups' interests precedence over the interests of the state as a whole.

The complex history of the former Yugoslavia, which was characterized by centuries of ethnic and religious struggle, can be used to pinpoint the origins of postsocialist nationalism in that region. But during the socialist rule of Josip Broz Tito, the nation had a time of comparatively good stability and prosperity because to a socialist federalist model that tried to balance the interests of many ethnic groups. However, this system started to fall apart in the 1980s as a result of political corruption, economic stagnation, and Tito's death in 1980, which left a power vacuum that was filled by nationalist forces.

The growth of neoliberalism, the end of the Cold War, and the collapse of the Soviet Union were some of the larger geopolitical trends that influenced the creation of postsocialist nationalism in the former Yugoslavia. These events produced a new international climate that favored the pursuit of economic and political sovereignty as well as the reassertion of national identities. This movement was most noticeable in the Balkans, where various ethnic groups started to stake out their claims to land, resources, and political influence.�

Postsocialist nationalism had a significant impact on the former Yugoslavia, and its legacy is still influencing the region today. A string of bloody wars, including the Bosnian-Herzegovina War and the Kosovo War, which claimed the lives of hundreds of thousands of people and forced millions to flee their homes, marked the breakup of Yugoslavia. Conflicting nationalist narratives that presented various ethnic groups as victims and defended violent acts as essential for self-defense were the driving forces behind these conflicts.

In the former Yugoslavia, the effects of postsocialist nationalism are still felt today in the form of persistent ethnic tensions, political division, and nationalist narratives. Promoting reconciliation, establishing democratic institutions, and cultivating a sense of a common national identity have been the main focuses of efforts to address these issues. The region continues to be a complex and contentious space, characterized by its complicated past and current battles to balance the conflicting aspirations of various ethnic groups, despite the

Figure B8: AIDT23-06-SBB

COMPARISON OUTPUT

FILE: AIDT23-06-TFO-orig.txt

Phishing is a type of online scam where criminals use email or other forms of communication to trick individuals into sharing sensitive information like passwords or financial data. Unfortunately, I had a personal experience with phishing emails and I want to share my story to help others avoid falling into the same trap.

It all started when I received an email from what appeared to be my bank, asking me to click on a link to verify my account information. The email looked legitimate, with the same logo and color scheme as my bank's website, and it even had a sense of urgency, saying that my account would be suspended if I didn't verify my information within 24 hours.

Without thinking twice, I clicked on the link and was taken to a website that looked exactly like my bank's website. I entered my login information and then was asked to fill out a form with my name, address, and social security number. Again, I didn't think twice and filled out the form.

It wasn't until later that day when I received another email from my bank, warning customers about a phishing scam that was going around. My heart sank as I realized that I had fallen for it. I immediately went to my bank's website and changed my password, but I couldn't shake the feeling that my personal information was now in the hands of criminals.

Luckily, I didn't suffer any financial loss, but the experience taught me a valuable lesson about the importance of being vigilant when it comes to online security. Here are a few tips that I learned from my experience that can help others avoid falling for phishing scams:

Always double-check the sender's email address. Scammers will often use a similar email address to the real company, but there will be a slight difference. Check the domain name to ensure it's the same as the real company.

FILE: AIDT23-06-TFO-edit.txt

Phishing is a type of online scam in which fraudsters coerce victims into disclosing sensitive information like passwords or financial information by using email or other forms of contact. I regretfully had a bad experience with scam emails, and I want to share it here to warn others and prevent them from making the same mistake.

Everything began when I opened an email that purported to be from my bank and asked me to click on a link to confirm my account details. The email had the same logo and color scheme as my bank's website and appeared to be genuine. It even had a sense of urgency, threatening to suspend my account if I didn't validate my information within 24 hours.

I clicked the link without giving it a second thought, and I was then directed to a website that resembled my bank's website precisely. After entering my login details, I was required to complete a form by providing my name, location, and social security number. Again, I completed the paperwork without hesitation.

I didn't receive another email from my bank alerting customers about a phishing scam until later that day. I was shocked to learn that I had been duped, and my spirit sank. I changed my password right away on the website for my bank, but I couldn't get rid of the feeling that my confidential information was now in the hands of thieves.

Fortunately, I didn't lose any money, but the experience made me realize how crucial it is to be vigilant when it comes to internet security. Here are a few pointers I discovered from my experience to help people stay away from hacking scams:

Always verify the email address of the sender. Scammers frequently use email addresses that are somewhat identical to those of the legitimate business. Verify the domain name to make sure it corresponds to the actual business.

Figure B9: AIDT23-06-TFO